%% file: main.tex
\documentclass[11pt,a4paper]{article}
\PassOptionsToPackage{table,dvipsnames,svgnames,x11names}{xcolor}

\usepackage[T1]{fontenc}
\usepackage[utf8]{inputenc}
\usepackage{times}

\usepackage[margin=2.4cm,headheight=38pt,headsep=18pt,footskip=28pt]{geometry}
\usepackage{setspace}
\usepackage[table,dvipsnames]{xcolor}
\usepackage{colortbl}
\usepackage{graphicx}
\usepackage{booktabs}
\usepackage{array}
\usepackage{multirow}
\usepackage{tabularx}

\usepackage{amsmath,amssymb}

\definecolor{AMapBlue}{HTML}{48B8F0}
\definecolor{AMapDeepBlue}{HTML}{0877C9}
\definecolor{AMapSky}{HTML}{EAF8FF}
\definecolor{AMapLine}{HTML}{9EDBFA}
\definecolor{TextBlack}{HTML}{20242A}

\usepackage[numbers,sort&compress]{natbib}

\usepackage{enumitem}
\setlist{nosep,leftmargin=1.5em}
\usepackage{caption}
\usepackage{fancyhdr}
\usepackage{lastpage}
\usepackage{titlesec}
\usepackage[skins,breakable]{tcolorbox}
\usepackage[colorlinks=true,linkcolor=AMapDeepBlue,citecolor=AMapDeepBlue,urlcolor=AMapDeepBlue]{hyperref}

\newcommand{\reportname}{GAD}
\newcommand{\reportlogo}{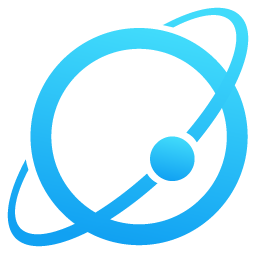}
\newcommand{\reportlogoheight}{18pt}

\newcommand{\orgname}{AMAP CV Lab \textbar{} Alibaba Group}
\newcommand{\reportcenter}[1]{\raisebox{\dimexpr(\depth-\height)/2\relax}{#1}}
\newlength{\reporttextbaselineoffset}
\newcommand{\reporttextalign}[1]{\raisebox{\reporttextbaselineoffset}{#1}}

\renewcommand{\headrulewidth}{1.2pt}
\renewcommand{\footrulewidth}{0.8pt}
\renewcommand{\headrule}{\hbox to\headwidth{\color{AMapBlue}\leaders\hrule height \headrulewidth\hfill}}
\renewcommand{\footrule}{\hbox to\headwidth{\color{AMapLine}\leaders\hrule height \footrulewidth\hfill}}

\titleformat{\section}
  {\Large\bfseries\color{AMapDeepBlue}}
  {\thesection}{0.75em}{}
  [\vspace{0.32em}{\color{AMapLine}\titlerule[0.6pt]}]
\titleformat{\subsection}
  {\large\bfseries\color{AMapDeepBlue}}
  {\thesubsection}{0.65em}{}
\titleformat{\subsubsection}
  {\normalsize\bfseries\color{AMapDeepBlue}}
  {\thesubsubsection}{0.55em}{}
\titlespacing*{\section}{0pt}{1.55em plus 0.25em minus 0.1em}{0.95em}
\titlespacing*{\subsection}{0pt}{1.25em plus 0.2em minus 0.1em}{0.55em}
\titlespacing*{\subsubsection}{0pt}{1.05em plus 0.15em minus 0.08em}{0.45em}

\newtcolorbox{abstractbox}{
  enhanced,
  breakable,
  colback=AMapSky,
  colframe=AMapBlue,
  coltitle=white,
  colbacktitle=AMapBlue,
  title=Abstract,
  fonttitle=\Large\bfseries,
  halign title=center,
  toptitle=5pt,
  bottomtitle=5pt,
  fontupper=\large,
  boxrule=0.8pt,
  arc=1.5mm,
  left=12pt,
  right=12pt,
  top=10pt,
  bottom=10pt,
  before skip=1.2em,
  after skip=1.4em
}

\definecolor{DimAction}{HTML}{0877C9}
\definecolor{DimVisual}{HTML}{2E9E5B}
\definecolor{DimPhysics}{HTML}{E07B00}
\definecolor{DimMemory}{HTML}{8E44AD}

\input{math_commands.tex}

\input{config.tex}
\author{
    \textbf{Junfeng Ni}$^{1,2,3,*}$ \quad
    \textbf{Ziyu Zhu}$^{2,*}$ \quad
    \textbf{Fan Jiang}$^{2,*,\dagger,\textrm{\normalfont \Letter}}$ \quad
    \textbf{Yu Liu}$^{1}$ \quad
    \textbf{Chiyu Wang}$^{2}$\\
    \textbf{Jiawei Zhang}$^{2}$ \quad
    \textbf{Zhaoxu Sun}$^{2}$ \quad
    \textbf{Mu Xu}$^{2}$ \quad
    \textbf{Siyuan Huang}$^{3,\textrm{\normalfont \Letter}}$ \quad
    \textbf{Yixin Chen}$^{3,\textrm{\normalfont \Letter}}$
    \vspace{5pt}\\
    \small $^{*}$Equal contribution \quad
    \small $^{\dagger}$Project leader \quad
    \small $^{\textrm{\normalfont \Letter}}$Corresponding author
    \vspace{2pt}\\
    \small $^{1}$Tsinghua University \quad
    \small $^{2}$AMAP, Alibaba Group \\
    \small $^{3}$State Key Laboratory of General Artificial Intelligence, BIGAI
}

\title{\modelname: Geometry Attention Distillation \\ for Consistent Video Generation}

\makeatletter
\renewcommand{\maketitle}{%
  \begin{center}
    {\bfseries\color{AMapDeepBlue}\LARGE \@title\par}
    \vspace{0.2em}
    {\large \@author\par}
  \end{center}
  \vspace{-1em}
}
\makeatother

\begin{document}
\thispagestyle{fancy}
\maketitle

\begin{abstractbox}
Video diffusion models generate visually compelling short-term sequences but often struggle to preserve scene geometry and object structure over time. 
In this paper, we first demonstrate that current video attention cannot identify reliable correspondences across frames, with responses dispersed across multiple regions even after recognizable scene structure has emerged. 
We argue that reliable cross-frame associations are essential for information exchange as scene structure forms in video generation. 
Thus, we introduce Geometry Attention Distillation (\modelname), which transfers correspondences from a frozen geometry model's attention to video attention.
Extracted from clean videos, these correspondences link the same static surface or moving object across frames. 
Since structure formation begins in deeper layers at higher noise and shallower layers at lower noise, we apply attention distillation at a student layer selected according to the noise timestep.
After training, video attention focuses more consistently on corresponding locations. 
Experiments across all three video backbones demonstrate improved long-range scene consistency and dynamic temporal consistency, with geometric gains across generation lengths.
On Wan2.2 5B, \modelname improves cross-frame matching accuracy by at least 30\% over the Wan baseline on both long-range benchmarks.
Notably, \modelname achieves these gains without additional inference modules, offering a promising approach to efficient and consistent video-based world simulation.
Project page: {\small\url{https://fantasy-amap.github.io/GAD/}}.
\end{abstractbox}

\begin{figure}[htbp]
    \centering
    \includegraphics[width=\linewidth]{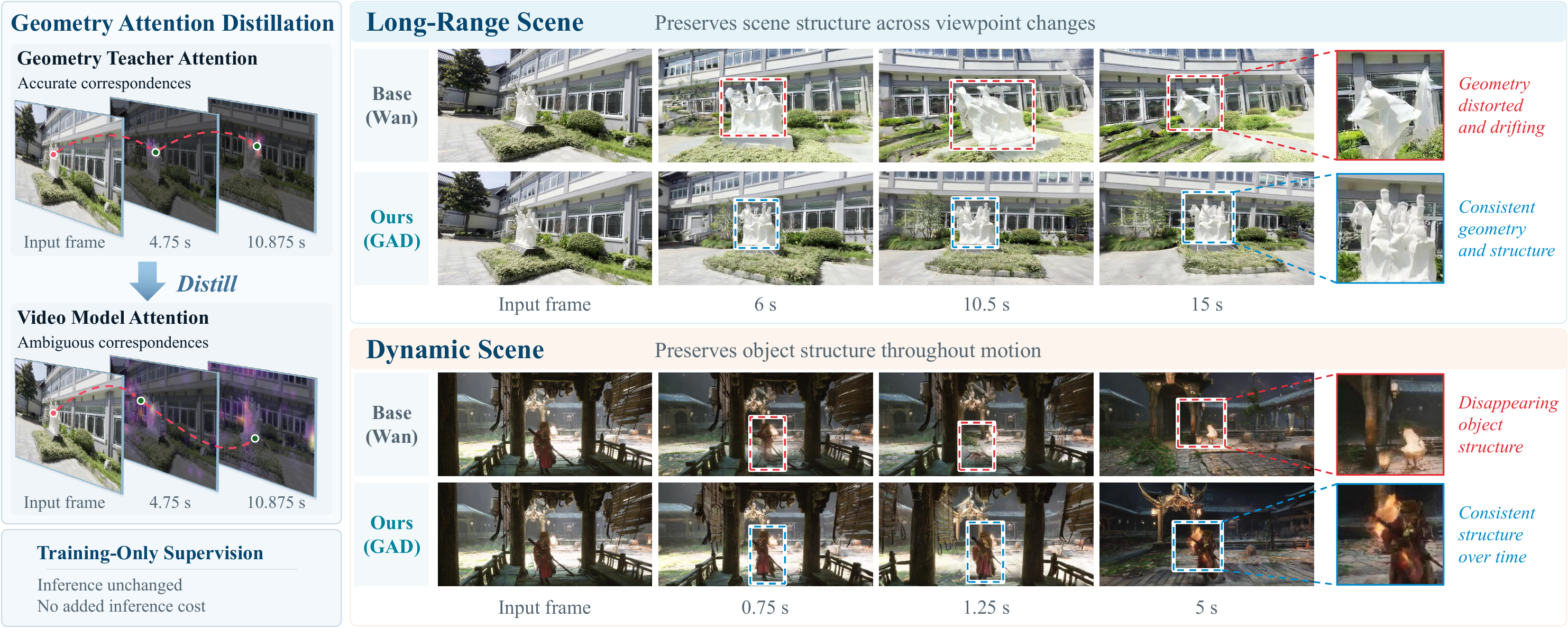}
    \caption{\textbf{Geometry attention distillation for consistent video generation.} Correspondences from a geometry teacher guide video attention during training (left), improving scene geometry under viewpoint changes and object structure during motion (right), without added inference cost.}
    \label{fig:teaser}
\end{figure}

\section{Introduction}
\label{sec:introduction}

Recent video diffusion models~\citep{blattmann2023stable,kong2024hunyuanvideo,yang2025cogvideox,wan2025wan} generate visually compelling sequences from text and images, yet struggle to preserve scene geometry over time and maintain coherent object structure during motion. Such consistency is essential for interactive world modeling~\citep{yang2023learning,parker2025genie}, robotic planning~\citep{du2023learning,ko2024learning}, and autonomous driving simulation~\citep{hu2023gaia,wang2024drivedreamer}, where generated videos must represent temporally coherent environments.

To improve consistency, existing approaches employ camera-aware positional encoding~\citep{li2026cameras,yin2026scope}, explicit geometric conditioning~\citep{ren2025gen3c,cao2025uni3c,huang2025voyager}, history-conditioned generation~\citep{li2025vmem,zhang2026frame}, and geometry-aware feature alignment~\citep{wu2026geometry,huang2026gen3r}. 
However, such guidance does not ensure that video models reliably associate observations of the same scene element across frames, a capability essential for consistent video generation.

To understand how cross-frame associations develop, we examine attention patterns in video models and the emergence of scene structure across noise levels and network depths.
Our analysis shows that video attention cannot reliably identify the same scene element across frames: its responses are often dispersed across multiple regions, even after recognizable scene structure has emerged. 
Such ambiguous associations can mix information from unrelated regions, undermining the preservation of scene geometry and object structure across frames.
Ideally, cross-frame attention should connect observations of the same scene element, enabling information exchange that supports consistent structure formation. We further observe that recognizable scene structure begins to emerge in deeper layers at higher noise and shallower layers at lower noise. This suggests guiding video attention where structure and cross-frame associations are beginning to form, with the guidance depth adapted to the noise level. For a source of this guidance, we turn to geometry foundation models~\citep{leroy2024grounding,wang2025vggt} with promising multi-view matching capabilities.
Probing VGGT-$\Omega$~\citep{wang2026vggtomega} reveals attention responses that associate the same static surface or moving object across frames, providing geometric correspondence targets for guiding video attention.

Building on these insights, we introduce Geometry Attention Distillation (\modelname), an attention distillation framework that guides cross-frame associations during structure formation. During training, we extract correspondence targets from the attention of a frozen geometry model on clean videos. The total attention each source token assigns to a target frame serves as a reliability cue for weighting its supervision. We select the student layer for attention distillation according to the noise timestep, guiding cross-frame associations as scene structure forms. At the selected layer, pairwise supervision encourages spatial discrimination within frame pairs, while trackwise supervision extends temporal coverage by matching a common source to multiple target frames. These complementary distillation objectives are optimized jointly with the original video generation loss, transferring cross-frame relations while preserving the flexibility of video representations. After training, video attention focuses more consistently on corresponding regions across frames, while generated videos better preserve scene geometry and moving-object structure.

We evaluate \modelname on camera-controlled video generation using RealEstate10K and DL3DV-10K for long-range scene consistency and OmniWorld-Game for dynamic consistency.
Across all three video backbones, \modelname improves geometric consistency in long-range generation and temporal consistency in dynamic generation.
Geometric improvements across generation lengths further support the applicability of \modelname across backbone configurations and video durations.
Notably, with Wan2.2 5B, \modelname delivers at least a 30\% relative improvement in cross-frame matching accuracy on both long-range benchmarks, without additional inference modules or computations.

Our main contributions are summarized as follows:
\begin{itemize}[leftmargin=*,itemsep=2pt,topsep=3pt]
  \item We analyze cross-frame associations in video models, revealing persistent geometric ambiguities in attention and a noise-dependent shift in the depth at which scene structure begins to form. These observations motivate geometry attention distillation at layers where structure is emerging.
  \item We propose \modelname, an attention distillation framework that transfers geometric correspondence knowledge to video attention for improved cross-frame consistency.
  \item Extensive experiments demonstrate that \modelname improves long-range scene consistency and dynamic temporal consistency across all three video backbones, with consistent improvements across generation lengths and no additional inference modules.
\end{itemize}

\section{Related Work}
\label{sec:related}

\subsection{Consistent Video Generation}
\label{sec:related-consistency}

Video diffusion models~\citep{ho2022video,blattmann2023stable,kong2024hunyuanvideo,yang2025cogvideox,wan2025wan} generate visually compelling videos but still struggle to maintain consistent scene geometry and object structure across frames. Existing efforts~\citep{yang2025longlive,huang2025selfforcing,zhu2026causal} address these limitations through temporal coordination, historical context, and spatial guidance. Sampling strategies improve temporal continuity~\citep{qiu2024freenoise,lu2024freelong,yang2026diff,chen2026freespec} while attention design regulates information exchange~\citep{zhou2024storydiffusion,tang2026temporal}. Long-range generation further benefits from compressed or hierarchical context~\citep{zhang2026frame,zhu2025memorize,zhao2026densitykv,chandratreya2026millivid}, selective memory access~\citep{wu2025pack,ji2025memflow,ding2026layerrecall,yang2026decmem}, and training that addresses error accumulation and long-context dependencies~\citep{yin2025slow,chang2026towards,chen2026context}. Camera-aware positional encodings complement this temporal context with spatial guidance~\citep{li2026cameras,yin2026scope,jang2026rays}. 
These approaches improve information retention and exchange across frames, but do not necessarily ensure that attention connects observations of the same scene element. Complementing these efforts, \modelname uses geometric correspondences to directly supervise video attention, guiding associations between observations of the same scene element across viewpoints and through motion.

\subsection{Geometry Foundation Models for Video Consistency}
\label{sec:related-geometry-foundations}

Geometry foundation models~\citep{wang2024dust3r,wang2025continuous,wang2025vggt} make learned structural priors accessible to video generation. Together with explicit representations, these priors provide spatial references for view-conditioned synthesis~\citep{li2025vmem,yu2025wonderworld,yu2024viewcrafter,liu2026reconx} and persistent scene memory~\citep{huang2025voyager,wang2026latent,lv2026octworld}. Geometry also guides learning through feature alignment and positional encoding~\citep{wu2026geometry,jin2026crepe}, joint appearance--geometry modeling~\citep{lu2025movis,dai2026fantasyworld,chen2025deepverse}, and depth or motion constraints~\citep{bai2026geovideo,fang2025worldreel}. Preference optimization and reinforcement learning further use geometric feedback to improve generated content~\citep{du2026videogpa,an2026vggrpo,wang2026world,ackermann2026geoflow}. However, methods using explicit geometric conditions~\citep{ren2025gen3c,cao2025uni3c} can inherit reconstruction and reprojection errors and face additional challenges in dynamic scenes. 
While implicit geometric learning~\citep{huang2026gen3r,hu2026geometry} offers greater flexibility for dynamic scenes, direct feature alignment may still impose representational constraints beyond those needed for geometric consistency.
To this end, \modelname draws correspondence targets from geometry model attention to supervise cross-frame associations in video attention, transferring relations between observations of the same scene element without matching feature representations or requiring reconstructed geometry at inference.

\subsection{Attention Distillation and Correspondence Supervision}
\label{sec:related-transfer}

Knowledge distillation can transfer attention patterns and relational structure in addition to individual features~\citep{zagoruyko2016paying,tung2019similarity,qiu2023sats}, with cross-stage connections shaping how teacher knowledge reaches the student~\citep{chen2021distilling}. In diffusion models, feature alignment~\citep{yu2024representation,hwang2025cross,bhowmik2026moalign} and token-relation distillation~\citep{zhang2026videorepa,lian2026sara} provide complementary supervision, refined through token selection or noise-dependent prior granularity~\citep{pang2026maskalign,min2026ahpa}. 
Diffusion representations also support correspondence estimation~\citep{tang2023emergent}, consistent video editing~\citep{geyer2024tokenflow,qi2023fatezero}, and sampling guidance~\citep{zu2026training}. 
More directly, CAMEO~\citep{kwon2026correspondence} improves novel-view synthesis by supervising a fixed attention layer with correspondences derived from static 3D pointmaps. 
For consistent video generation, \modelname instead extracts reliability-weighted correspondences from geometry model attention across static surfaces and moving object parts, and routes attention distillation to student layers according to the noise timestep to guide cross-frame associations during structure formation.

\begin{figure}[t!]
  \centering
  \begin{subfigure}[b]{\linewidth}
    \captionsetup{font=scriptsize,skip=2pt}
    \centering
    \includegraphics[width=\linewidth]{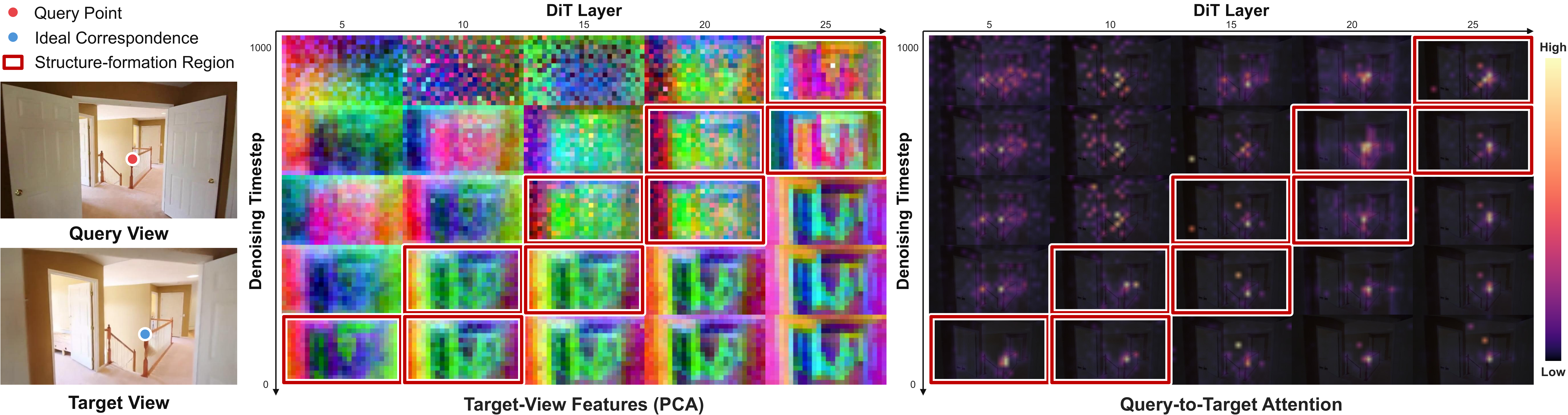}
    \subcaption{Features and attention from video generative models across noise levels and layers.}
    \label{fig:noise-depth}
  \end{subfigure}

  \begin{subfigure}[b]{\linewidth}
    \captionsetup{font=scriptsize,skip=2pt}
    \centering
    \includegraphics[width=\linewidth]{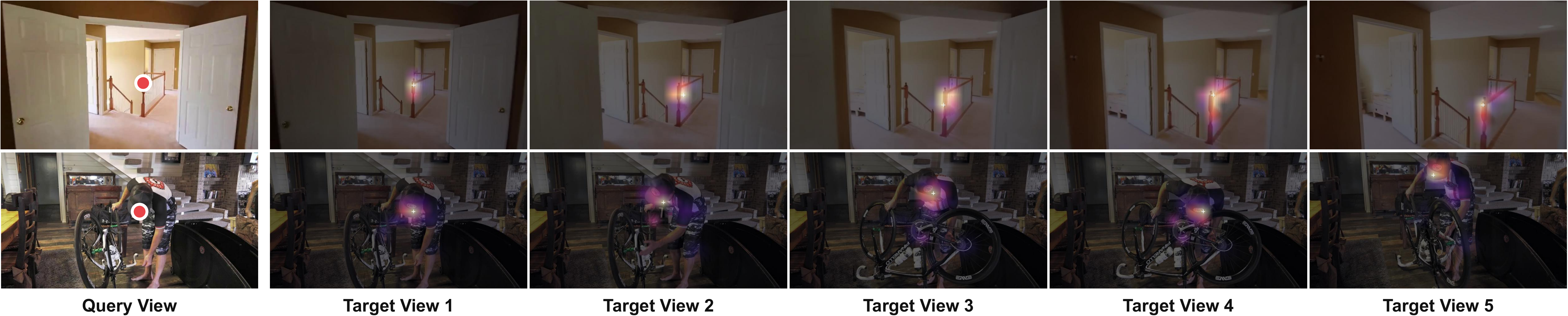}
    \subcaption{Attention from VGGT-$\Omega$ on static surfaces and moving objects.}
    \label{fig:vggt-omega}
  \end{subfigure}
  \vspace{-0.25in}
  \caption{\textbf{Correspondence patterns in video generation and geometry foundation models.} (a) In the \textit{structure-formation region} (red boxes), recognizable scene structure and cross-frame associations begin to emerge, at deeper layers under higher noise and shallower layers under lower noise. However, attention remains ambiguous even after structure becomes clear. (b) Selected VGGT-$\Omega$ heads track the same static surface or moving object across frames.}
  \label{fig:attention-analysis}
\end{figure}

\section{Analysis and Motivation}
\label{sec:motivation}

\subsection{Attention Patterns in Video Models}
\label{sec:video-attention}
\label{sec:noise-depth}

Cross-frame consistency requires associating the same scene element across viewpoint changes or object motion. We examine how video attention establishes these associations across noise levels and network depths.
In~\cref{fig:noise-depth}, we fix a source query and a target frame within a single video generation trajectory, and visualize the target frame's intermediate token features using PCA alongside query-to-target attention responses. The PCA visualizations reveal recognizable scene structure, while the attention maps show whether responses concentrate near the reference correspondence or favor other regions. Further visualization details are provided in~\cref{app:video-attention}.

This joint analysis reveals two key findings. First, video attention cannot reliably identify the same scene element across frames. Its responses can remain dispersed across multiple regions, including those with similar appearance or semantics, even after recognizable scene structure has emerged in the PCA features. Second, the emergence of scene structure and cross-frame associations depends on both noise level and network depth. We distinguish three qualitative regimes. 
In the \textit{noise-dominated region}, PCA features show little recognizable scene structure, while attention responses remain scattered rather than concentrated around the corresponding region. 
In the \textit{structure-formation region} (red boxes in~\cref{fig:noise-depth}), recognizable scene structure begins to emerge, and attention starts to respond to corresponding regions across frames, although competing responses remain. This region occurs in deeper layers at higher noise and shallower layers at lower noise.
In the \textit{structure-established region}, scene layout is more clearly defined in the PCA features and cross-frame associations are more apparent, but they are not necessarily geometrically accurate. 

These observations motivate explicitly guiding video attention toward geometrically corresponding regions to help preserve scene and object structure across frames. 
Furthermore, the three observed regimes suggest that the placement of this guidance matters. In the noise-dominated region, weak structural evidence may make correspondence supervision difficult to exploit; in the structure-established region, additional guidance may have less opportunity to shape associations that have already formed. 
The structure-formation region offers an intermediate stage where correspondence learning has a structural basis while the associations are still developing. We therefore hypothesize that geometric guidance is most effective in this region. Since this region emerges at different depths depending on noise, we adapt the supervision layer to the noise timestep.

\subsection{Geometric Correspondence in Geometry Foundation Models}
\label{sec:geometry-attention}

Recent advances in geometry foundation models have demonstrated strong multi-view matching capabilities, motivating their use as a source of correspondence supervision. Probing VGGT-$\Omega$~\citep{wang2026vggtomega}, we find that attention from selected layers and heads captures correspondences on both static surfaces and moving objects. As shown in~\cref{fig:vggt-omega}, attention responses remain concentrated around the same railing location across viewpoint changes in the static example, and follow the person's hat as the person moves in the dynamic example. These observations suggest that geometry model attention can guide associations with the same scene element across both camera and object motion. 
Visualization details and layer--head selection are described in~\cref{app:geometry-attention,app:teacher-details}.

However, VGGT-$\Omega$ can still produce an attention peak when the scene element represented by a source token is occluded or absent from the target view, making the peak alone an unreliable correspondence cue. As illustrated in~\cref{fig:more_vggt_omega}, ambiguous or incorrect responses can exhibit visible peaks while assigning little total attention to the target frame's spatial tokens. We call this total the target-frame attention mass and use it as a reliability cue to downweight potentially unreliable matches.

\begin{figure}[t]
  \centering
  \includegraphics[width=\linewidth]{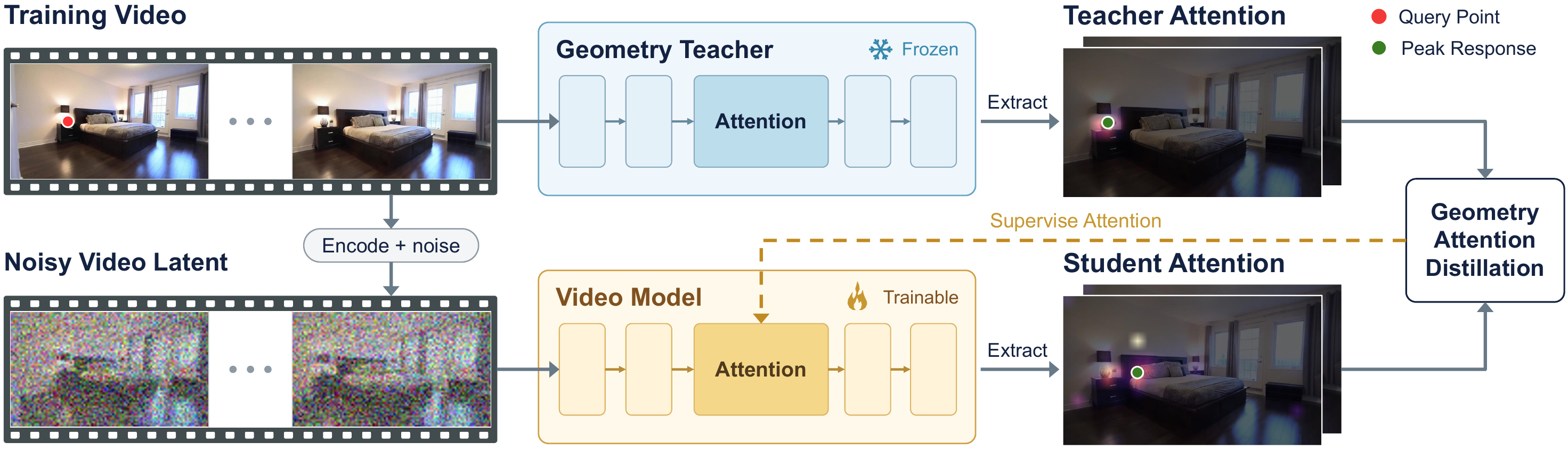}
  \caption{\textbf{Overview of \modelname.} Correspondences extracted by a frozen geometry teacher from clean videos guide the video model's attention during denoising. The correspondence supervision complements the generation objective during training, without changing the inference procedure.}
  \label{fig:method}
\end{figure}

\section{Geometry Attention Distillation}
\label{sec:method}

Building on the analysis in~\cref{sec:motivation}, we distill geometric correspondences from VGGT-$\Omega$ attention into video model attention to improve cross-frame consistency, as illustrated in~\cref{fig:method}.

\subsection{Confidence-Aware Teacher Correspondences}
\label{sec:teacher}

Given a clean training video, we extract attention from the frozen VGGT-$\Omega$ teacher and align its responses with the student's spatial and temporal token grid. Teacher attention is normalized jointly across all input frames, allowing the total attention assigned to each target frame to serve as a reliability cue, as discussed in~\cref{sec:geometry-attention}.
Let $a=(i,u)$ denote source token $u$ in frame $i$, and let $\bar A_{a,j}(v)$ denote its aligned attention weight at spatial token $v$ in target frame $j\ne i$, retaining this original frame mass. We extract the candidate correspondence and target-frame attention mass as
\begin{equation}
  c_{a,j}=\arg\max_v\bar A_{a,j}(v),
  \qquad m_{a,j}=\sum_v\bar A_{a,j}(v).
  \label{eq:teacher}
\end{equation}
We convert the attention mass into a reliability weight $w_{a,j}=f(m_{a,j})$, where $f$ is a nondecreasing mapping to $[0,1]$ and invalid relations receive zero weight. 

To tolerate small localization errors, we construct a soft target $y_{a,j}(v)$ by Gaussian smoothing around $c_{a,j}$. It is normalized over the valid spatial neighborhood $\mathcal N(c_{a,j})$ and zero elsewhere in the target frame. Thus, $y_{a,j}$ specifies where the student should match, while $w_{a,j}$ controls its relative contribution to the supervision. The target preserves the teacher-derived correspondence rather than reproducing its full attention distribution. Teacher attention alignment, confidence weighting, and target smoothing are detailed in~\cref{app:teacher-details}. 

\subsection{Student Attention Distillation with Noise-Timestep Routing}
\label{sec:student}

\paragraph{Noise-timestep routing.}
To guide cross-frame associations during structure formation, we adapt the supervision depth to the noise level, following the analysis in~\cref{sec:noise-depth}. Let $t\in[0,1000]$ denote the model noise timestep, with larger $t$ indicating higher noise. We select one student layer per training sample using interval boundaries $0=b_0<\cdots<b_K=1000$:
\begin{equation}
  \ell(t)=l_k \quad\text{for}\quad b_{k-1}\le t<b_k,
  \qquad k=1,\ldots,K,
  \label{eq:routing}
\end{equation}
where $K$ is the number of noise intervals, $l_k$ is the layer assigned to interval $k$, and the final interval includes $t=1000$. 
Specifically, we assign shallower layers to lower noise and deeper layers to higher noise to follow the noise-dependent onset of structure formation.

\paragraph{Correspondence supervision.}
At the selected layer $\ell(t)$, we aggregate query--key scores across attention heads to form a student matching distribution. Inspired by CAMEO~\citep{kwon2026correspondence}, a lightweight head-aggregation MLP weights and combines these scores, followed by a softmax over candidate tokens $\mathcal K$ in target frame $j$. For source token $a$, we denote the resulting distribution by $P_{a,j}(v\mid\mathcal K)$, omitting the layer index for brevity. This aggregation is used only during training.

For each source--target-frame relation $(a,j)$, the candidate set contains the full positive neighborhood, $\mathcal N(c_{a,j})\subseteq\mathcal K$. We compare the student distribution with the teacher-derived target through a per-correspondence cross-entropy loss:
\begin{equation}
  \mathcal L_{\mathrm{corr}}(a,j;\mathcal K)=-\sum_{v\in\mathcal N(c_{a,j})}
  y_{a,j}(v)\log P_{a,j}(v\mid\mathcal K).
  \label{eq:matching}
\end{equation}
When aggregating these per-correspondence losses, we use reliability weight $w_{a,j}$ to control the relative contribution of each correspondence.

To select the correspondences used for supervision, we employ two complementary sampling schemes that balance spatial discrimination and temporal coverage.
\textit{Pairwise supervision} samples source tokens from frame pairs and compares each with all spatial tokens in its target frame, providing a full set of spatial alternatives for matching. \textit{Trackwise supervision} uses correspondences from a common source token to multiple target frames, using the positive neighborhood and sampled negative tokens in each frame to extend temporal coverage at lower cost than evaluating all spatial candidates. The corresponding objectives, $\mathcal L_{\mathrm{pair}}$ and $\mathcal L_{\mathrm{track}}$, aggregate the per-correspondence losses $\mathcal L_{\mathrm{corr}}$ using their reliability weights. Attention aggregation and routing configurations are detailed in~\cref{app:student-details}, while sampling procedures and loss reductions are specified in~\cref{app:loss}.

\subsection{Overall Training Procedure}
\label{sec:training}

We retain the flow-matching objective~\citep{lipman2022flow} and add correspondence losses:
\begin{equation}
  \mathcal L_{\mathrm{total}}=\mathcal L_{\mathrm{FM}}
  +\lambda_1\mathcal L_{\mathrm{pair}}
  +\lambda_2\mathcal L_{\mathrm{track}},
  \label{eq:overall}
\end{equation}
where $\lambda_1$ and $\lambda_2$ control the pairwise and trackwise supervision strengths, respectively. Both correspondence losses are evaluated at the routed layer $\ell(t)$, with averaging over training samples and noise levels implicit. 
The flow-matching objective continues to train the backbone's denoising velocity prediction, while the correspondence losses guide cross-frame associations.

In practice, we first cache teacher correspondences and confidence information, preserving alignment with the student's temporal sampling and spatial transformations. Each iteration constructs the noisy video latent $z_t$ at noise timestep $t$, selects layer $\ell(t)$, and computes the pairwise and trackwise losses from its query--key responses. We update the video transformer and aggregation MLPs while keeping the geometry teacher and latent encoder frozen. 
At inference, only the trained video model is used, with no additional inference modules or correspondence caches; its attention computation and sampling procedure remain unchanged.
The flow-matching formulation and training configurations are detailed in~\cref{app:training-details}.

\section{Experiments}
\label{sec:experiments}

We evaluate \modelname on camera-controlled video generation, assessing long-range scene consistency and dynamic temporal consistency alongside generation quality and inference efficiency. 
Implementation and evaluation details are provided in~\cref{app:implementation,app:protocol}, while limitations are discussed in~\cref{app:limitations}.

\subsection{Setting}
\label{sec:setup}

\paragraph{Training.}
Our training corpus comprises 35K clips from RealEstate10K~\citep{zhou2018stereo}, DL3DV-10K~\citep{ling2024dl3dv}, and OmniWorld-Game (\ie, OmniWorld's game subset~\citep{zhou2025omniworld}), covering real-world scenes and game environments. We train at a resolution of $832\times480$ for 10 epochs on 32 NVIDIA A800 GPUs. We fine-tune the video DiT and aggregation MLPs with AdamW at a learning rate of $10^{-5}$ while keeping the text encoder and VAE frozen.

\begin{table}[t]
  \caption{\textbf{Long-range camera-controlled video generation on RealEstate10K and DL3DV-10K.} \modelname improves all seven metrics over the Wan baselines across all three backbones. Top-3 results are highlighted as the \colorbox{gbest}{first}, \colorbox{gsecond}{second} and \colorbox{gthird}{third}.}
  \label{tab:static}
  \centering
  \begingroup
  \scriptsize
  \setlength{\tabcolsep}{2.5pt}
  \renewcommand{\arraystretch}{1.15}
  \resizebox{\linewidth}{!}{%
  \begin{tabular}{@{}l*{7}{c}@{\hspace{7pt}}*{7}{c}@{}}
    \toprule
    \multirow{2}{*}{Method} & \multicolumn{7}{c}{RealEstate10K} & \multicolumn{7}{c}{DL3DV-10K} \\
    \cmidrule(lr){2-8}\cmidrule(lr){9-15}
    & FID$\downarrow$ & FVD$\downarrow$ & CLIP$\uparrow$ & $R_{\text{dist}}\downarrow$ & $T_{\text{dist}}\downarrow$ & $M_{\text{rate}}\uparrow$ & $M_{\text{dist}}\downarrow$
    & FID$\downarrow$ & FVD$\downarrow$ & CLIP$\uparrow$ & $R_{\text{dist}}\downarrow$ & $T_{\text{dist}}\downarrow$ & $M_{\text{rate}}\uparrow$ & $M_{\text{dist}}\downarrow$ \\
    \midrule
    GF & 61.46 & 464.8 & 30.03 & 6.229 & 0.134 & 0.188 & 55.66 & 104.1 & 813.7 & 29.42 & 8.145 & 0.126 & 0.209 & 54.50 \\
    Gen3R & 33.35 & 305.3 & 30.88 & 5.167 & 0.103 & 0.321 & 36.52 & 45.23 & 373.7 & 32.66 & 6.457 & 0.088 & 0.310 & 39.27 \\
    Voyager & 58.38 & 410.3 & 30.47 & 4.779 & 0.099 & 0.334 & 33.01 & 65.77 & 608.4 & 32.16 & \cellcolor{gthird}4.740 & 0.073 & 0.313 & 41.02 \\
    Gen3C & 31.07 & 264.7 & 31.11 & \cellcolor{gsecond}4.288 & \cellcolor{gsecond}0.076 & 0.315 & 35.07 & 38.63 & 325.6 & 33.46 & \cellcolor{gsecond}4.732 & \cellcolor{gsecond}0.056 & 0.289 & 42.19 \\
    Uni3C & 34.46 & 289.6 & 31.32 & 5.146 & 0.096 & 0.384 & 32.11 & 43.28 & 351.1 & 33.30 & 5.589 & 0.078 & 0.331 & \cellcolor{gthird}38.42 \\
    \midrule
    Wan2.1\,1.3B & 30.83 & 267.4 & 31.11 & 7.236 & 0.162 & 0.336 & 38.19 & 40.06 & 336.9 & 33.93 & 6.395 & 0.112 & 0.318 & 42.41 \\
    \quad $+$\,\modelname & 30.53 & 259.0 & 31.69 & 4.734 & 0.093 & \cellcolor{gthird}0.387 & 32.16 & 38.83 & 328.8 & \cellcolor{gthird}34.13 & 5.029 & 0.071 & \cellcolor{gthird}0.358 & \cellcolor{gsecond}37.05 \\
    \midrule
    Wan2.2\,5B & 31.32 & 258.8 & 31.77 & 6.705 & 0.175 & 0.263 & 45.53 & 38.60 & 322.1 & 33.85 & 9.573 & 0.108 & 0.257 & 48.90 \\
    \quad $+$\,\modelname & \cellcolor{gthird}29.20 & \cellcolor{gthird}242.1 & \cellcolor{gsecond}32.03 & \cellcolor{gthird}4.335 & 0.088 & 0.350 & \cellcolor{gthird}31.36 & \cellcolor{gthird}37.93 & \cellcolor{gthird}318.0 & \cellcolor{gsecond}34.24 & 4.890 & \cellcolor{gthird}0.065 & 0.335 & 38.85 \\
    \midrule
    Wan2.1\,14B & \cellcolor{gsecond}24.80 & \cellcolor{gsecond}201.8 & \cellcolor{gthird}32.00 & 4.724 & \cellcolor{gthird}0.078 & \cellcolor{gsecond}0.413 & \cellcolor{gsecond}30.09 & \cellcolor{gsecond}31.02 & \cellcolor{gsecond}255.8 & 34.05 & 5.497 & 0.083 & \cellcolor{gsecond}0.365 & 39.68 \\
    \quad $+$\,\modelname & \cellcolor{gbest}23.15 & \cellcolor{gbest}194.8 & \cellcolor{gbest}32.14 & \cellcolor{gbest}2.754 & \cellcolor{gbest}0.058 & \cellcolor{gbest}0.476 & \cellcolor{gbest}26.01 & \cellcolor{gbest}28.66 & \cellcolor{gbest}228.8 & \cellcolor{gbest}34.34 & \cellcolor{gbest}3.039 & \cellcolor{gbest}0.046 & \cellcolor{gbest}0.393 & \cellcolor{gbest}34.10 \\
    \bottomrule
  \end{tabular}%
  }
  \endgroup
\end{table}

\begin{table}[t]
  \caption{\textbf{Dynamic camera-controlled video generation on OmniWorld-Game.} \modelname improves subject and background consistency and motion smoothness, without added inference cost.}
  \label{tab:dynamic}
  \centering
  \begingroup
  \scriptsize
  \setlength{\tabcolsep}{2.5pt}
  \renewcommand{\arraystretch}{1.15}
  \resizebox{\linewidth}{!}{%
  \begin{tabular}{@{}l*{3}{c}@{\hspace{7pt}}*{5}{c}@{\hspace{7pt}}*{3}{c}@{}}
    \toprule
    \multirow{2}{*}{Method} & \multicolumn{3}{c}{Rendering} & \multicolumn{5}{c}{VBench Metrics} & \multicolumn{3}{c}{Inference Cost} \\
    \cmidrule(lr){2-4}\cmidrule(lr){5-9}\cmidrule(lr){10-12}
    & FID$\downarrow$ & FVD$\downarrow$ & CLIP$\uparrow$
    & Subj. Cons.$\uparrow$ & Bg. Cons.$\uparrow$ & Motion Sm.$\uparrow$ & Aesth. Q.$\uparrow$ & Image Q.$\uparrow$
    & Time (s)$\downarrow$ & Peak Mem. (GB)$\downarrow$ & Avg. Mem. (GB)$\downarrow$ \\
    \midrule
    GF & 162.0 & 1543 & 20.48 & 0.698 & 0.851 & 0.961 & 0.397 & 0.380 & 194.6 & \cellcolor{gthird}48.96 & 33.80 \\
    Gen3R & 81.91 & 870.1 & 26.84 & 0.817 & 0.907 & 0.957 & 0.482 & 0.505 & 209.3 & 50.66 & 37.05 \\
    Voyager & 103.6 & 987.7 & 26.09 & 0.805 & 0.906 & 0.963 & 0.492 & 0.521 & 2583.2 & 67.45 & 62.27 \\
    Gen3C & 70.41 & 774.8 & 27.55 & 0.810 & 0.908 & 0.970 & 0.493 & 0.527 & 825.9 & 63.71 & 57.20 \\
    Uni3C & 80.63 & 856.6 & 27.56 & 0.804 & 0.908 & 0.968 & 0.504 & 0.553 & 444.0 & 49.77 & 37.68 \\
    \midrule
    GT & - & - & - & 0.867 & 0.919 & 0.981 & 0.553 & 0.662 & - & - & - \\
    \midrule
    Wan2.1\,1.3B & 65.16 & 665.9 & 27.39 & 0.813 & 0.889 & 0.968 & 0.533 & 0.591 & 172.9 & \cellcolor{gbest}25.44 & \cellcolor{gbest}22.23 \\
    \quad $+$\,\modelname & 60.35 & 625.7 & 27.67 & \cellcolor{gsecond}0.854 & \cellcolor{gthird}0.910 & \cellcolor{gsecond}0.977 & 0.534 & 0.586 & \cellcolor{gthird}171.2 & \cellcolor{gbest}25.44 & \cellcolor{gsecond}22.76 \\
    \midrule
    Wan2.2\,5B & 61.29 & 657.6 & 27.46 & 0.825 & 0.900 & 0.964 & \cellcolor{gthird}0.542 & \cellcolor{gsecond}0.636 & \cellcolor{gbest}69.60 & \cellcolor{gsecond}34.85 & 26.11 \\
    \quad $+$\,\modelname & \cellcolor{gthird}59.52 & \cellcolor{gthird}622.3 & \cellcolor{gsecond}28.23 & \cellcolor{gthird}0.853 & \cellcolor{gsecond}0.913 & \cellcolor{gthird}0.976 & \cellcolor{gsecond}0.546 & \cellcolor{gbest}0.641 & \cellcolor{gsecond}70.44 & \cellcolor{gsecond}34.85 & \cellcolor{gthird}26.08 \\
    \midrule
    Wan2.1\,14B & \cellcolor{gsecond}45.94 & \cellcolor{gsecond}419.7 & \cellcolor{gthird}28.18 & 0.850 & 0.909 & 0.971 & \cellcolor{gsecond}0.546 & 0.612 & 854.9 & 57.54 & 55.74 \\
    \quad $+$\,\modelname & \cellcolor{gbest}44.75 & \cellcolor{gbest}410.2 & \cellcolor{gbest}28.88 & \cellcolor{gbest}0.863 & \cellcolor{gbest}0.917 & \cellcolor{gbest}0.978 & \cellcolor{gbest}0.552 & \cellcolor{gthird}0.617 & 856.5 & 57.54 & 55.48 \\
    \bottomrule
  \end{tabular}%
  }
  \endgroup
\end{table}

\paragraph{Datasets.}
All three benchmarks evaluate camera-controlled video generation conditioned on a first frame, a text prompt, and a target camera trajectory. We assess long-range scene consistency with 15-second videos on RealEstate10K and DL3DV-10K, and dynamic temporal consistency with 5-second videos on OmniWorld-Game. 
For all three benchmarks, we use scene-disjoint training and test splits to prevent data leakage between training and evaluation.

\paragraph{Baselines.}
We compare with Geometry Forcing (GF)~\citep{wu2026geometry} and Gen3R~\citep{huang2026gen3r}, which incorporate geometry through learned representations, and Voyager~\citep{huang2025voyager}, Gen3C~\citep{ren2025gen3c}, and Uni3C~\citep{cao2025uni3c}, which use explicit geometric conditioning. We apply \modelname to three video backbones: Wan2.1 1.3B, Wan2.2 5B, and Wan2.1 14B~\citep{wan2025wan}. For each backbone, the Wan baseline is fine-tuned on the same training data using only the flow-matching objective, under training settings matched to \modelname. Throughout our experimental comparisons, Wan denotes this fine-tuned baseline, and $+\,\modelname$ denotes training with the additional distillation objectives. Baseline adaptation procedures are detailed in~\cref{app:baseline-adaptation}.

\paragraph{Metrics.}
We report FID, FVD, and CLIP for generation quality and text alignment. To evaluate geometric consistency on RealEstate10K and DL3DV-10K, we measure camera accuracy using rotation distance $R_{\text{dist}}$ (degrees) and normalized translation error $T_{\text{dist}}$, following~\citet{li2025vmem}, and cross-view consistency using matching hit rate $M_{\text{rate}}$ and mean target-position error $M_{\text{dist}}$ (pixels), following~\citet{lu2025movis}. For dynamic scenes on OmniWorld-Game, we report five VBench metrics~\citep{huang2024vbench}: subject consistency, background consistency, motion smoothness, aesthetic quality, and imaging quality, with ground-truth scores provided for reference. Finally, we assess inference efficiency through runtime and GPU memory usage measured on an NVIDIA A800 GPU.

\begin{table}[t]
  \centering
  \begin{minipage}[t]{0.382679\linewidth}
    \centering
    \captionsetup{justification=raggedright,singlelinecheck=false}
    \caption{\textbf{Generation duration.} \modelname improves geometry at each duration.}
    \label{tab:duration}
    \scriptsize
    \setlength{\tabcolsep}{3pt}
    \renewcommand{\arraystretch}{1.186229}
    \resizebox{\linewidth}{!}{%
  \begin{tabular}{@{}l*{4}{c}@{}}
    \toprule
    Duration & $R_{\text{dist}}\downarrow$ & $T_{\text{dist}}\downarrow$ & $M_{\text{rate}}\uparrow$ & $M_{\text{dist}}\downarrow$ \\
    \midrule
    5s & 3.020 & 0.064 & 0.529 & 21.84 \\
    5s $+$\,\modelname & \textbf{1.684} & \textbf{0.053} & \textbf{0.571} & \textbf{19.31} \\
    \midrule
    10s & 5.948 & 0.072 & 0.371 & 38.34 \\
    10s $+$\,\modelname & \textbf{3.376} & \textbf{0.056} & \textbf{0.453} & \textbf{33.33} \\
    \midrule
    15s & 9.573 & 0.108 & 0.257 & 48.90 \\
    15s $+$\,\modelname & \textbf{4.890} & \textbf{0.065} & \textbf{0.335} & \textbf{38.85} \\
    \bottomrule
  \end{tabular}%
    }
  \end{minipage}\hfill
  \begin{minipage}[t]{0.597321\linewidth}
    \centering
    \captionsetup{justification=raggedright,singlelinecheck=false}
    \caption{\textbf{Noise-timestep routing ablation.} Our proposed schedule (R8) yields the best geometric scores among the tested schedules.}
    \label{tab:timestep-routing}
    \scriptsize
    \setlength{\tabcolsep}{3pt}
    \renewcommand{\arraystretch}{1.15}
    \resizebox{\linewidth}{!}{%
  \begin{tabular}{@{}ll*{4}{c}@{}}
    \toprule
    Exp. ID & Loss Assignment & $R_{\text{dist}}\downarrow$ & $T_{\text{dist}}\downarrow$ & $M_{\text{rate}}\uparrow$ & $M_{\text{dist}}\downarrow$ \\
    \midrule
    R1 (Wan) & None & 9.573 & 0.108 & 0.257 & 48.90 \\
    \midrule
    R2 & L5 at All noise & 7.440 & 0.086 & 0.295 & 43.89 \\
    R3 & L15 at All noise & 7.346 & 0.090 & 0.286 & 42.71 \\
    R4 & L25 at All noise & 6.540 & 0.075 & 0.289 & 43.55 \\
    \midrule
    R5 & L5 only at High noise & 8.932 & 0.098 & 0.262 & 46.36 \\
    R6 & L25 only at Low noise & 9.290 & 0.106 & 0.258 & 47.17 \\
    R7 & L25 / L5 at Low / High noise & 8.755 & 0.104 & 0.268 & 45.77 \\
    \midrule
    \textbf{R8 (Ours)} & L5 / L15 / L25 at Low / Medium / High noise & \textbf{4.890} & \textbf{0.065} & \textbf{0.335} & \textbf{38.85} \\
    \bottomrule
  \end{tabular}%
    }
  \end{minipage}
\end{table}

\begin{figure}[t]
  \centering
  \includegraphics[width=\linewidth]{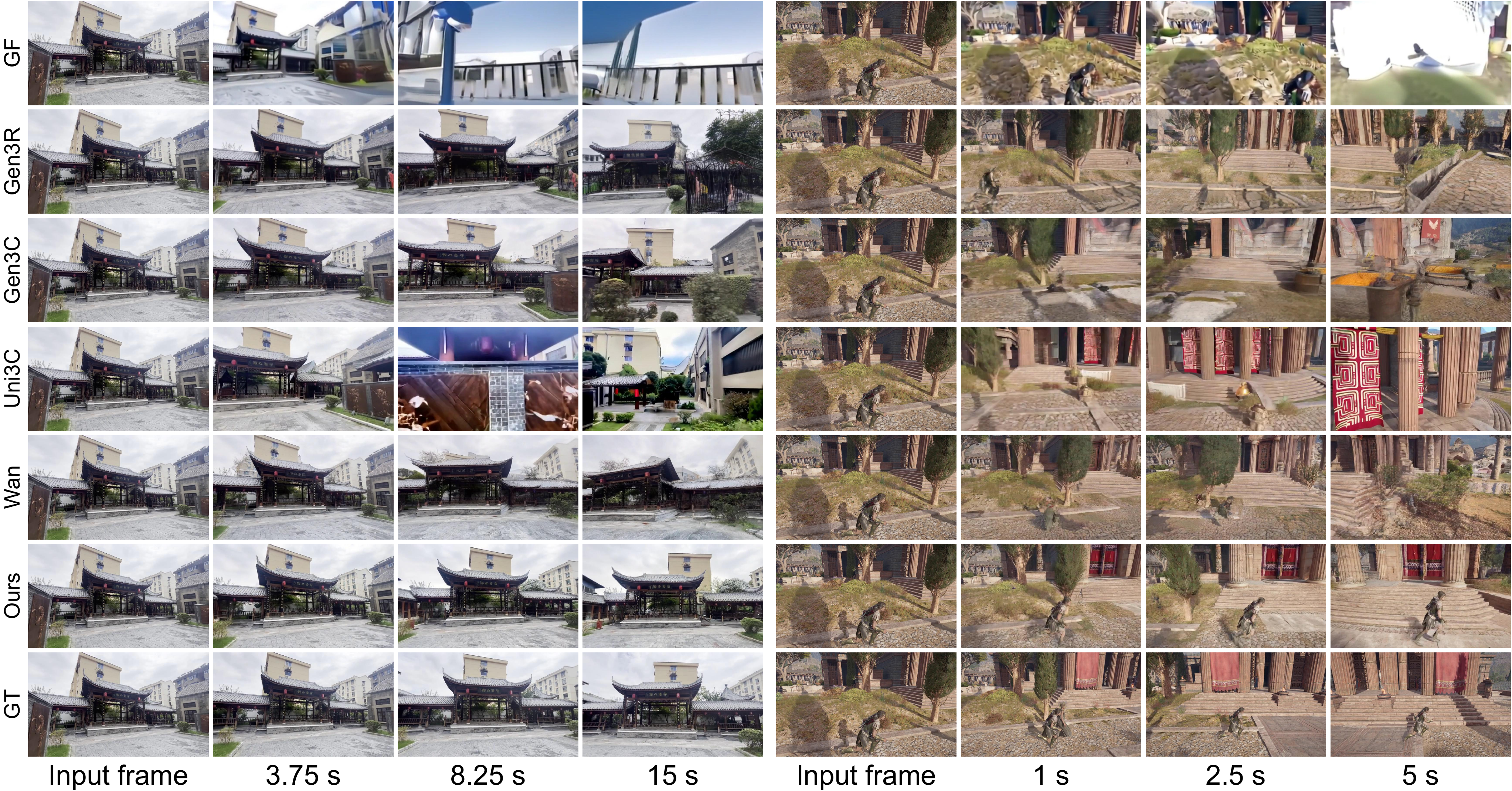}
  \caption{\textbf{Qualitative comparison of camera-controlled video generation.} \modelname better preserves scene geometry over 15 seconds (left) and moving-object structure over 5 seconds (right).}
  \label{fig:qualitative}
\end{figure}

\subsection{Results}
\label{sec:main-results}

\paragraph{Long-range scene consistency.}
Compared with the Wan baselines, \modelname improves cross-view matching and camera accuracy while achieving better FID, FVD, and CLIP scores across both datasets and all three backbones~\cref{tab:static}. Notably, on Wan2.2 5B, matching hit rate increases by at least 30\% on each dataset. These gains demonstrate improved geometric consistency alongside better generation quality.
With Wan2.1 14B, \modelname achieves the best results across all seven metrics on both datasets, indicating that its benefits extend to a larger backbone.
Qualitative comparisons in~\cref{fig:qualitative,fig:qual_static_results_s1} further show that \modelname better preserves scene layout and geometric structure throughout long-range camera-controlled sequences.

\paragraph{Dynamic video consistency.}
Across all three backbones, \modelname improves subject consistency, background consistency, and motion smoothness over the corresponding Wan baseline (\cref{tab:dynamic}). With Wan2.1 14B, it achieves the best scores on these metrics among the evaluated methods. Improvements in both subject and background scores suggest that correspondence supervision is useful beyond static scene content, consistent with transferring relations between moving objects rather than enforcing fixed spatial locations. Lower FID and FVD and higher CLIP scores further indicate that these gains coexist with improved generation quality and text alignment. Qualitative comparisons in~\cref{fig:qualitative,fig:qual_dynamic_results_s2} complement these scores: \modelname better preserves the structure of moving characters and foreground objects, while maintaining a more stable background layout as the viewpoint changes. 
Attention visualizations in~\cref{fig:attention_comparison} further show that, compared with the Wan baseline, \modelname focuses video attention more consistently on corresponding scene regions in both static and dynamic examples. These more localized responses accompany better preservation of scene geometry and moving-object structure in the generated videos.

\paragraph{Applicability and inference efficiency.}
On DL3DV-10K with Wan2.2 5B, we further evaluate generation at 5, 10, and 15 seconds, separately training \modelname and its Wan baseline under matched settings for each duration. As shown in~\cref{tab:duration}, \modelname improves all four geometric metrics at every duration, and complete results in~\cref{tab:duration-full} show broadly comparable or improved generation quality. Together, these results demonstrate the applicability of our method across backbone configurations and generation lengths.
Notably, \modelname achieves these gains without additional inference modules or computations, preserving the backbone's inference procedure. 
\cref{tab:dynamic} reports unchanged peak GPU memory and minor runtime differences consistent with run-to-run variability. Time-weighted average memory is also sensitive to variations in the duration of individual execution stages.

\begin{table}[t]
  \caption{\textbf{Distillation component ablation.} Confidence-aware teacher weighting (CAT) combined with trackwise supervision (TWS) and pairwise supervision (PWS) gives the best results across all seven metrics.}
  \label{tab:ablation}
  \centering
  \begingroup
  \scriptsize
  \setlength{\tabcolsep}{4pt}
  \renewcommand{\arraystretch}{1.15}
  \begin{tabular}{@{}ccc@{\hspace{10pt}}*{7}{c}@{}}
    \toprule
    CAT & TWS & PWS & FID$\downarrow$ & FVD$\downarrow$ & CLIP$\uparrow$ & $R_{\text{dist}}\downarrow$ & $T_{\text{dist}}\downarrow$ & $M_{\text{rate}}\uparrow$ & $M_{\text{dist}}\downarrow$ \\
    \midrule
    $\times$ & $\times$ & $\times$ & 38.60 & 322.1 & 33.85 & 9.573 & 0.108 & 0.257 & 48.90 \\
    $\times$ & $\checkmark$ & $\checkmark$ & 39.04 & 329.4 & 33.92 & 8.753 & 0.092 & 0.230 & 42.94 \\
    $\checkmark$ & $\checkmark$ & $\times$ & 38.61 & 321.4 & 34.01 & 6.028 & 0.082 & 0.274 & 40.20 \\
    $\checkmark$ & $\times$ & $\checkmark$ & 38.35 & 320.3 & 34.05 & 5.312 & 0.076 & 0.291 & 39.67 \\
    $\checkmark$ & $\checkmark$ & $\checkmark$ & \textbf{37.93} & \textbf{318.0} & \textbf{34.24} & \textbf{4.890} & \textbf{0.065} & \textbf{0.335} & \textbf{38.85} \\
    \bottomrule
  \end{tabular}%
  \endgroup
\end{table}

\subsection{Ablations}
\label{sec:ablation}

Using Wan2.2 5B on DL3DV-10K, we ablate the noise-timestep routing schedule, confidence-aware teacher weighting, and pairwise and trackwise supervision to assess their contributions.

\paragraph{Noise-dependent supervision.}
As shown in~\cref{tab:timestep-routing}, fixed-layer supervision (R2--R4) already improves geometric metrics over the Wan baseline (R1), while the proposed noise-timestep routing (R8) outperforms all other configurations (R1--R7); complete metrics appear in~\cref{tab:timestep-routing-full}. Thus, correspondence targets are useful on their own, while their placement determines how effectively they guide generation. The lower performance of R5--R7 compared with R2--R4 further suggests that varying supervision with noise is insufficient without an appropriate layer choice. 
These results support the motivation in~\cref{sec:motivation}: supervision should adapt to the noise timestep to target the structure-formation region rather than constrain a fixed layer throughout denoising.

\paragraph{Contributions of distillation components.}
\cref{tab:ablation} shows that transferring teacher correspondences indiscriminately is insufficient: without confidence weighting, matching hit rate falls below the Wan baseline despite modest improvements in camera error. This supports accounting for uncertainty in teacher matches rather than treating every target equally. With confidence weighting, either pairwise or trackwise supervision improves all geometric metrics, while their combination performs best across all seven metrics. The joint gains are consistent with their complementary roles: pairwise supervision discriminates spatial matches within a target frame, whereas trackwise supervision extends correspondence coverage across time.

\section{Conclusion}
\label{sec:conclusion}

We have presented \modelname, an attention distillation framework that transfers geometric correspondence knowledge from a frozen geometry teacher to video attention. 
Motivated by geometric ambiguities in video attention and the noise-dependent emergence of scene structure, \modelname guides cross-frame associations during structure formation while preserving the video model's representational flexibility.
Experiments show that \modelname consistently focuses attention on corresponding regions and improves long-range scene consistency and dynamic temporal consistency across all three video backbones, with geometric improvements across generation lengths. These gains require no additional inference modules. Our findings highlight geometric correspondence transfer through attention as a promising approach to consistent video generation and video-based world simulation.

\clearpage
\subsection*{AI Use Statement}
AI tools assisted with revising prose and theoretical exposition, LaTeX formatting, and scripts to assemble and visualize analysis outputs. The authors designed the method and experiments, ran the experiments, verified the reported results and mathematical arguments, and take full responsibility for the paper.

\subsection*{Reproducibility Statement}
\cref{sec:method} defines the teacher targets, student responses, routing rule, and training objective. Sampling procedures and loss reductions are detailed in~\cref{app:loss}, while implementation and evaluation protocols are provided in~\cref{app:implementation,app:protocol}. Complete routing and duration results appear in~\cref{app:additional-ablations}, and the PCA and attention analysis protocols for~\cref{fig:attention-analysis} are described in~\cref{app:pca-attention}.

\bibliography{iclr2027_conference}
\bibliographystyle{unsrtnat}

\clearpage
\appendix
\section{Loss and Sampling Details}
\label[appendix]{app:loss}

\paragraph{Sampling source--target relations.}
Pairwise supervision samples valid directed frame pairs without replacement, excluding pairs within the same frame. Within each selected pair, source tokens are sampled uniformly without replacement from those with valid cached correspondences. Each selected query is compared with all spatial keys in its target frame. If fewer valid pairs or queries are available than the sampling budget, the remaining entries are masked out.

Trackwise supervision first samples one source frame containing eligible source tokens, then samples tokens that have valid correspondences in at least two other frames. Each source is matched to all of its valid target frames, excluding its own frame. For each target frame, the candidate set contains the Gaussian positive neighborhood and uniformly sampled spatial negatives without replacement from outside that neighborhood. Both schemes resample their relations during training; confidence weights affect the loss reduction rather than the probability of selecting a valid relation.

\paragraph{Pairwise reduction.}
Let $\mathcal P$ denote the sampled valid source--target-frame relations, and $\Omega_j$ all spatial tokens in target frame $j$. Using $\mathcal L_{\mathrm{corr}}$ from~\cref{eq:matching}, the pairwise loss is
\begin{equation}
 \mathcal L_{\mathrm{pair}}=
 \frac{\sum_{(a,j)\in\mathcal P}w_{a,j}
 \left[\mathcal L_{\mathrm{corr}}(a,j;\Omega_j)+\eta \mathcal L_{\mathrm{hard}}(a,j)\right]}
 {\sum_{(a,j)\in\mathcal P}w_{a,j}}.
 \label{eq:pair-reduction}
\end{equation}
The hard-negative term contrasts the weighted positive partition with the highest-scoring negatives outside the positive neighborhood:
\begin{equation}
 \begin{aligned}
 Z^+_{a,j}&=\sum_{v\in\mathcal N(c_{a,j})}y_{a,j}(v)\exp s_{a,j}(v),\\
 Z^-_{a,j}&=\sum_{v\in\operatorname{Top}_{K_{\mathrm{neg}}}(\Omega_j\setminus\mathcal N(c_{a,j}))}
             \exp s_{a,j}(v),\\
 \mathcal L_{\mathrm{hard}}(a,j)&=-\log\frac{Z^+_{a,j}}{Z^+_{a,j}+Z^-_{a,j}}.
 \end{aligned}
 \label{eq:hard}
\end{equation}
Top-$K_{\mathrm{neg}}$ selection uses the student logits, with $K_{\mathrm{neg}}=64$. The positive target is normalized over valid in-grid neighbors; invalid border positions are excluded. The hard-negative coefficient $\eta$ is fixed within each controlled comparison.

\paragraph{Trackwise reduction.}
Let $\mathcal S$ be sampled source tokens and $\mathcal J_a$ the valid target frames for source $a$, excluding its own frame. Each $\mathcal K_{a,j}$ contains the positive neighborhood and randomly sampled negatives from outside it. We define
\begin{equation}
 \begin{aligned}
 \overline{\mathcal L}_{\mathrm{corr},a}&=\frac{\sum_{j\in\mathcal J_a}w_{a,j}\mathcal L_{\mathrm{corr}}(a,j;\mathcal K_{a,j})}
                        {\sum_{j\in\mathcal J_a}w_{a,j}},\\
 \overline w_a&=\frac{1}{|\mathcal J_a|}\sum_{j\in\mathcal J_a}w_{a,j},\\
 \mathcal L_{\mathrm{track}}&=
 \frac{\sum_{a\in\mathcal S}\overline w_a\overline{\mathcal L}_{\mathrm{corr},a}}
      {\sum_{a\in\mathcal S}\overline w_a}.
 \end{aligned}
 \label{eq:track-reduction}
\end{equation}
With uniform confidence, each source contributes its average target-frame loss, irrespective of track length. With nonuniform confidence, uncertain target frames contribute less within a source, and sources with low average confidence contribute less overall. Empty valid sets produce zero auxiliary loss; implementations clamp denominators for numerical stability.

\paragraph{Combining the two objectives.}
The implementation uses an overall auxiliary coefficient $\lambda_{\mathrm{aux}}$ and relative branch weights $\beta_{\mathrm{pair}}$ and $\beta_{\mathrm{track}}$. With both branches active, the coefficients in~\cref{eq:overall} are
\begin{equation}
 \begin{aligned}
 \lambda_1&=\lambda_{\mathrm{aux}}
 \frac{\beta_{\mathrm{pair}}}{\beta_{\mathrm{pair}}+\beta_{\mathrm{track}}},\\
 \lambda_2&=\lambda_{\mathrm{aux}}
 \frac{\beta_{\mathrm{track}}}{\beta_{\mathrm{pair}}+\beta_{\mathrm{track}}}.
 \end{aligned}
\end{equation}
When only one branch is enabled or has valid supervision, the branch weights are renormalized over the active branch. If neither branch has valid supervision, the auxiliary loss is zero. This convention keeps the overall auxiliary coefficient separate from the allocation between the two objectives.

\section{Implementation Details}
\label[appendix]{app:implementation}

\subsection{Teacher Targets and Alignment}
\label[appendix]{app:teacher-details}
\paragraph{Reference correspondences for head selection.}
We select the teacher layer--head pair using reference correspondences from both static and dynamic scenes. For the static set, we randomly sample 50 DL3DV scenes and estimate per-view depth and camera parameters with VGGT-$\Omega$, retaining only high-confidence depth estimates. Inspired by CAMEO~\citep{kwon2026correspondence}, we construct correspondence labels through cycle projection. A sampled source pixel is unprojected into 3D using its depth and camera parameters, then projected into a target view and assigned to a target token. We unproject the center of that target token using the target-view depth and project it back into the source view. A correspondence is retained only if the projections remain in bounds, the projected and estimated depths agree in both directions, and the return location is sufficiently close to the source sample. When multiple samples within a source token pass these checks, we retain the correspondence with the smallest cycle error. These are geometry-derived reference labels, rather than manually annotated ground truth. For the dynamic set, we use the provided ground-truth point correspondences from 30 dynamic scenes in  TAPVid-DAVIS~\citep{doersch2022tap}.

\paragraph{Layer--head evaluation and selection.}
We randomly sample query tokens from the input frame of each video and examine every global-attention layer--head pair. For each query, we extract the maximum-response location in each subsequent view and its target-frame attention mass, following~\cref{eq:teacher}. We compare the predicted locations with the reference correspondences to measure correspondence error and examine the associated attention masses as reliability cues. Considering correspondence consistency across both static and dynamic tokens, we select layer 13, head 10, which provides the best overall performance in this evaluation. All layer and head indices here are zero-based, and this single pair is used for training-label extraction in all experiments.

The evaluation also reveals substantial specialization across heads. For example, layer 13, head 0 captures static-token correspondences well but fails to reliably associate dynamic tokens, whereas layer 12, head 12 captures dynamic-token correspondences well but performs poorly on static tokens. Other pairs exhibit little correspondence consistency for either type of token. These observations motivate selecting a pair using both static and dynamic scenes, rather than relying on either set alone.

\paragraph{Limitations of projection-based labels.}
The depth-and-camera cycle projection above assumes a static scene and therefore cannot directly recover correspondences for moving scene elements. We also observe substantial errors around thin structures and closely spaced surfaces, where depth errors and token discretization can associate different surfaces. These limitations can affect methods that rely on such projection-derived supervision, including CAMEO. Extracting correspondences from VGGT-$\Omega$ attention instead exploits the model's learned cross-view consistency without imposing a static-scene projection constraint. In our observations, this alleviates these failure cases and also supports correspondences for dynamic scene elements.

\paragraph{Teacher extraction.}
We precompute correspondence targets with the frozen VGGT-$\Omega$ 1B teacher, using its 512-resolution checkpoint and global-attention layer 13, head 10 (zero-based indices). The teacher processes clean video frames. Attention is normalized over all tokens in all teacher input frames, including special tokens, whereas the target-frame mass sums only the patch-token responses in that frame.

Source patch queries are resampled to the student spatial grid before query normalization. For each target frame, the conditional patch-attention map is spatially resampled and renormalized before taking its maximum; its frame attention mass is retained separately. In~\cref{eq:teacher}, $\bar A$ denotes this aligned conditional map multiplied by the retained mass. Its sum and argmax therefore recover the cached mass and correspondence, respectively. The cache stores these quantities rather than the full attention matrix.

\paragraph{Temporal and spatial alignment.}
Teacher inputs follow the temporal representative indices used by the training label pipeline. For a $4n+1$-frame clip, these are $[0,2,6,\ldots,4n-2]$, with the clip's source offset applied when reading the original video. The resulting $n+1$ teacher views align with the student's latent time axis. The spatial grids are $30\times52$ for Wan2.1 1.3B and 14B, and $15\times26$ for Wan2.2 5B at $832\times480$ resolution. RGB frames, camera intrinsics, and teacher targets use consistent crop and resize geometry. Cache metadata records the source frames, clip offset, grid, and teacher identity, and is checked when loading labels. When a shorter prefix is read from a longer cache, the retained masses remain those computed in the original teacher context; they are not recomputed by renormalizing over the prefix.

\paragraph{Confidence weighting.}
We map the target-frame mass to a continuous weight:
\begin{equation}
  w_{a,j}=\chi_{a,j}
  \left[\sigmoid\left(\frac{\log_2\!\left(\max(m_{a,j},\varepsilon)\right)-\mu}{\kappa}\right)\right]^\gamma,
  \label{eq:confidence-map}
\end{equation}
where $\chi_{a,j}\in\{0,1\}$ indicates a valid cached relation. We use $\varepsilon=2^{-20}$ and set $\gamma=2$ empirically to further downweight low-confidence relations. The grid-specific $(\mu,\kappa)$ values are approximately $(-9.86777,1.81857)$ for the $30\times52$ grid and $(-9.75580,1.59004)$ for the $15\times26$ grid. The logarithm accommodates attention masses spanning multiple orders of magnitude, and the lower bound avoids taking the logarithm of zero. These weights set the relative contributions of correspondences in~\cref{eq:pair-reduction,eq:track-reduction}.

\paragraph{Human visibility annotations.}
We determine the sigmoid mapping from human visibility annotations on DL3DV, using attention labels computed in an 81-frame teacher context. The annotated subset contains 55 scenes: 48 randomly sampled scenes and seven manually selected challenging scenes involving thin structures, occlusions, or out-of-view content. Each annotation page shows one source token and seven target frames sampled from seven temporal intervals, independently of attention mass. Annotators judge whether the source-token content is visible, not visible, or uncertain in each target frame; attention statistics are hidden during annotation. Of 803 judgments, we exclude 230 uncertain cases and retain 573 source-token/target-frame pairs, comprising 401 visible and 172 not-visible cases. These labels assess visibility rather than the localization accuracy of the attention maximum.

\paragraph{Selecting and fitting the mass mapping.}
We compare three attention statistics as visibility cues for each source-token/target-frame pair $(a,j)$. The \emph{mass}, $m_{a,j}=\sum_{v\in\Omega_j}\bar A_{a,j}(v)$, is the total attention assigned to the target frame's spatial patch tokens, with the original normalization over all tokens in all teacher input frames preserved, as in~\cref{eq:teacher}. To describe how this attention is distributed within the target frame, let $p_{a,j}(v)=\bar A_{a,j}(v)/m_{a,j}$ denote the conditional distribution on the aligned spatial grid $\Omega_j$ for $m_{a,j}>0$. The \emph{normalized attention entropy} is $H_{a,j}=-\sum_{v\in\Omega_j}p_{a,j}(v)\log p_{a,j}(v)/\log|\Omega_j|$, with $0\log0=0$; it ranges from zero for a single-token peak to one for a uniform distribution. The \emph{top-1 attention response} is $\max_{v\in\Omega_j}p_{a,j}(v)$, the largest conditional attention probability in that frame. Thus, mass measures how much attention reaches a frame, whereas entropy and top-1 describe its spatial concentration within that frame. We use $\log_2(\max(m_{a,j},\varepsilon))$, $1-H_{a,j}$, and the top-1 response as regression features.

In five-fold cross-validation grouped by scene, logistic regression using only log-mass achieves ROC-AUCs of 0.903 and 0.932 for the $30\times52$ and $15\times26$ layouts, respectively. Adding entropy and the top-1 response increases these values only to 0.910 and 0.933, so we retain the simpler mass-only mapping. Grouping by scene keeps all pairs from a scene in the same fold.

For each layout, we fit a class-balanced logistic regression on all 573 definite annotations, with $x=\log_2(\max(m,\varepsilon))$ as its sole input. Its output $\sigmoid(ax+b)$ is equivalently written as $\sigmoid((x-\mu)/\kappa)$, with $\mu=-b/a$ and $\kappa=1/a$, yielding the parameters reported above. Class balancing reduces the influence of the larger visible class, yielding a balanced visibility confidence for weighting correspondence supervision. We use this continuous confidence instead of a hard visibility threshold, then apply the empirical exponent $\gamma$ in~\cref{eq:confidence-map}; the exponent is not part of the logistic fit. The mapping fitted on static DL3DV scenes is shared across static and dynamic supervision.

\paragraph{Local correspondence targets.}
We use a $3\times3$ Gaussian positive neighborhood with $\sigma=1$ in token units. Weights are computed from token-grid distances to the cached correspondence, with out-of-grid positions excluded and the remaining weights normalized to sum to one. This local target tolerates small localization errors without requiring the student to reproduce the teacher's full attention distribution.

\subsection{Student Responses and Noise-Timestep Routing}
\label[appendix]{app:student-details}
\paragraph{Multi-head response aggregation.}
Let $\mathbf q_a^{\ell,h}$ and $\mathbf k_{j,v}^{\ell,h}$ be the query and key after normalization and rotary positional encoding at layer $\ell$ and head $h$. We aggregate their relation scores with a learnable gate:
\begin{equation}
 \begin{aligned}
 r^h_{a,j}(v)&=\frac{\langle\mathbf q_a^{\ell,h},\mathbf k_{j,v}^{\ell,h}\rangle}{\sqrt d},\\
 \boldsymbol\alpha_{a,j}(v)&=\softmax\!\left(g_\phi(\mathbf r_{a,j}(v))\right),\\
 s_{a,j}(v)&=\frac{1}{\tau}\sum_{h=1}^{H}\alpha^h_{a,j}(v)r^h_{a,j}(v),\\
 P_{a,j}(v\mid\mathcal K)&=\frac{\exp s_{a,j}(v)}{\sum_{v'\in\mathcal K}\exp s_{a,j}(v')}.
 \end{aligned}
 \label{eq:student}
\end{equation}
Here, $d$ is the head dimension, $H$ is the number of heads, and $\mathcal K$ contains candidate spatial keys in the target frame. The head-aggregation MLP $g_\phi$ serves as the gate and has an $H\!\rightarrow\!4H\!\rightarrow\!H$ architecture with a GELU activation. Its final linear layer is initialized to zero, producing uniform head weights at initialization. Each supervised layer has its own gate. Auxiliary logits and aggregation are computed in FP32, with temperature $\tau=1$. The gate acts on each query--key pair and does not replace the attention operation used for generation. The pairwise hard-negative term uses $K_{\mathrm{neg}}=64$ negatives and coefficient $\eta=0.1$.

\paragraph{Noise coordinate and layer selection.}
Throughout the paper, $t\in[0,1000]$ denotes the model noise timestep presented to the network by the Wan scheduler after applying its noise schedule. The corresponding normalized interpolation coefficient is $s=t/1000\in[0,1]$. Noise-timestep routing compares this model noise timestep with the configured thresholds, rather than using the integer index of a training or inference step. We use the same routing schedule across all three backbones in both the static and dynamic generation settings: L5 for $t<850$, L15 for $850\le t<950$, and L25 for $t\ge950$, as specified in~\cref{tab:timestep-routing}. All layer indices are zero-based. Boundary values belong to the higher-noise interval, and only one layer contributes auxiliary supervision for a sample. The noise-timestep values in~\cref{fig:noise-depth} and the routing tables use this same coordinate.

\subsection{Training and Inference}
\label[appendix]{app:training-details}
\paragraph{Training configuration.}
As described in~\cref{sec:setup}, the training corpus contains 35K clips, and training runs for 10 epochs on 32 NVIDIA A800 GPUs. We fine-tune the video DiT and aggregation MLPs using AdamW with a learning rate of $10^{-5}$, keeping the text encoder, VAE, and geometry teacher frozen. Clips have a resolution of $832\times480$, with 241 frames for long-range generation and 81 frames for dynamic generation. The duration comparison additionally uses 161-frame clips. At 16 fps, the first-to-last-frame spans are 5, 10, and 15 seconds for 81, 161, and 241 frames. Each task and duration uses a separate Wan/\modelname training pair, sharing initialization, data, optimizer, learning-rate schedule, effective batch size, and training duration.

For joint pairwise and trackwise supervision, we sample up to 32 directed frame pairs with 256 source queries per pair, together with 256 trackwise source tokens and 128 random negatives per source--target-frame relation. In the single-branch ablations in~\cref{tab:ablation}, pairwise-only supervision uses up to 128 frame pairs with 320 queries per pair, while trackwise-only supervision uses up to 384 source tokens and 192 random negatives per source--target-frame relation. These budgets apply to each training sample at its selected supervision layer. Pairwise queries are compared with all spatial tokens in the target frame; trackwise sources are sampled from one source frame and matched to all valid target frames, with the positive neighborhood included in addition to the sampled negatives. Sampling and masking follow~\cref{app:loss}.

Across all \modelname experiments, including all three backbones and the static and dynamic generation settings, we use $\lambda_{\mathrm{aux}}=0.05$ and equal relative branch weights $\beta_{\mathrm{pair}}=\beta_{\mathrm{track}}=1$. Following the normalization in~\cref{app:loss}, this gives $\lambda_1=\lambda_2=0.025$ in~\cref{eq:overall} when both branches are active. If only one branch is active, its coefficient is $0.05$.

\paragraph{Flow-matching objective.}
For model noise timestep $t\in[0,1000]$, let $s=t/1000$ be the normalized interpolation coefficient. We sample $\epsilon\sim\mathcal N(0,I)$ and construct $z_t=(1-s)z_0+s\epsilon$. Under conditions $\mathbf c$, the video model $v_\theta$ receives $t$ and predicts the velocity $\epsilon-z_0$ with respect to $s$. We retain the backbone's noise-timestep sampling and loss weighting $\rho(t)$, expressed in the model-timestep coordinate:
\begin{equation}
  \mathcal L_{\mathrm{FM}}=\mathbb E_{z_0,\epsilon,t}
  \left[\rho(t)\|v_\theta(z_t,t,\mathbf c)-(\epsilon-z_0)\|_2^2\right].
  \label{eq:flow-matching}
\end{equation}
For all three benchmarks, $\mathbf c$ includes the first frame, a text prompt, and a target camera trajectory. The backbone's first-frame conditioning mechanism is retained during training and inference.

\paragraph{Auxiliary computation and inference.}
Pairwise supervision evaluates sampled queries against all spatial keys in their selected target frames. Trackwise supervision evaluates only the positive and sampled-negative candidates for its selected sources. Neither branch constructs a full all-frame attention matrix. These computations add training cost but are absent at inference, together with the teacher and correspondence caches. Inference uses only the trained video model and its original sampler; all requested frames are jointly denoised over the sampling steps. Paired Wan and \modelname evaluations use the same input conditions and generation seeds.

\section{Experimental Protocol}
\label[appendix]{app:protocol}

\subsection{Data Preparation and Conditioning}
We use RealEstate10K and DL3DV-10K for 15-second long-range generation and OmniWorld-Game for 5-second dynamic generation. DL3DV-10K uses its official evaluation set; RealEstate10K and OmniWorld-Game use scene-disjoint training and test splits. All methods receive the first image, the text prompt, and the target camera trajectory for the same clip.

RealEstate10K cameras are obtained from COLMAP reconstructions and aligned to the video frame indices. Their translation scale is aligned to a Depth Anything 3~\citep{lin2025da3} metric reference. DL3DV-10K uses VIPE camera estimates aligned to the extracted RGB frames. For OmniWorld-Game, we use the game subset from the xhigh tier of the SolarWM~\citep{huang2026solarwm} release, together with its camera annotations and captions. Spatial preprocessing applies the same crop and resize to RGB images and camera intrinsics. Camera trajectories are expressed relative to their first frame, and model-specific translation scaling is applied separately from the common evaluation trajectory.

\subsection{Baseline Adaptation}
\label[appendix]{app:baseline-adaptation}
The Wan baselines follow~\cref{app:implementation}. For GF, Gen3R, Voyager, Gen3C, and Uni3C, we use their released implementations and checkpoints with adapters for the shared image, prompt, and camera inputs. Each method retains its native generation procedure, including any depth estimation, geometry rendering, and history conditioning. Outputs are aligned to the common evaluation resolution of $832\times480$ and to 241 or 81 timestamps at 16 fps. Native frame counts and model resolutions can differ; the adapters transform camera intrinsics with the corresponding image geometry.

\paragraph{Long-range generation.}
GF retains its sparse-keyframe prediction and iterative interpolation procedure over the requested trajectory. Gen3R generates three 49-frame segments, conditioning each subsequent segment on the previous generated endpoint. Removing duplicate endpoints yields 145 native frames, which are aligned to the 241 evaluation timestamps by nearest-timestamp resampling. Voyager and Uni3C generate three 81-frame segments with one shared endpoint, yielding $81+80+80=241$ frames. Voyager propagates the previous generated RGB-D endpoint while retaining its initial world cache; its four internal prefix frames are removed from each segment's decoded output. Uni3C uses the previous generated endpoint as the next reference image and constructs its geometric conditions for the next segment. Gen3C uses its native long-video generation path over the requested camera sequence. Its adapter preserves the input field of view by resizing and padding to the native canvas, then removes the padding and resizes the output for evaluation. Segment continuation uses generated content rather than additional ground-truth RGB frames.

\paragraph{Dynamic generation.}
The OmniWorld-Game adapters cover the same 0--5-second interval. Voyager and Uni3C use one 81-frame segment, with Voyager retaining its internal prefix mechanism. GF retains its keyframe and interpolation procedure. Gen3R generates 49 native frames and resamples them to 81 output timestamps. Gen3C generates a 121-frame native window over the same interval and selects the nearest native timestamps for the 81-frame evaluation sequence. These conversions change the temporal sampling of the output, not the evaluated camera interval.

\paragraph{Camera-scale adaptation.}
Warping-based methods are sensitive to camera scale because their explicit scene initialization and geometric conditioning depend on depth estimated from the first frame. Camera translations must therefore be expressed at a scale consistent with this depth; a mismatch produces an inconsistent initial scene and incorrect warps, which can propagate errors into subsequent predictions. This dependence on depth--camera scale calibration is a practical limitation of these explicit geometric methods. To provide suitable geometric inputs for a fair comparison, we carefully align camera translations with the depth scale used by each baseline, using the dataset- and model-specific corrections described below. For Gen3C and Voyager, the DL3DV adapter applies the recorded VIPE-to-MoGe scale, approximately $0.370913$. On RealEstate10K, their input translations use the per-scene COLMAP-to-DA3 scale followed by the fixed DA3-to-MoGe factor, approximately $0.632710$. GF and Gen3R retain their internal camera normalization after the common input conversion. For Uni3C on RealEstate10K, camera translations are aligned to the DA3 depth scale using a per-scene scale factor.

For OmniWorld-Game, the supplied metric poses are scaled by approximately $0.227987$ for Gen3C/\allowbreak Voyager and $0.597926$ for Uni3C. These factors were estimated from training-side depth comparisons and fixed before evaluation. GF and Gen3R receive the metric poses without an additional dataset-specific multiplier. Scale factors affect translations only; rotations and intrinsics are unchanged. Common evaluation poses are stored separately and do not include model-specific scale factors.

\subsection{Evaluation Metrics}
\paragraph{Generation quality.}
We exclude the first conditioning frame from the target-frame quality measurements. FID uses 2048-dimensional Inception features from frames $4,8,\ldots,T-1$, pooled across the complete evaluation set. FVD uses 400-dimensional I3D features from the target-frame video clips and is also computed at the complete-set level. CLIP uses ViT-B/32 image and text embeddings, sampling every fourth target frame and averaging $\max(100\cos(\mathbf e_{\mathrm{image}},\mathbf e_{\mathrm{text}}),0)$.

\paragraph{Camera accuracy.}
Depth Anything 3~\citep{lin2025da3} estimates camera poses from the generated video. Predicted and reference trajectories are each transformed relative to their first camera, and their translations are independently normalized by the maximum distance from that camera. We average the rotation geodesic distance and normalized translation L2 distance over frames. Rotation distances are reported in degrees, as in~\cref{tab:static}; translation distances are dimensionless. This normalization evaluates relative trajectory agreement rather than absolute metric scale.

\paragraph{Cross-view consistency.}
LoMa-B~\citep{nordstrom2026loma} extracts up to 2,048 keypoints per frame. We compare first-frame-to-target-frame matches in generated and reference videos at frames $4,8,\ldots,T-1$. Reference matches are associated with generated matches by their source positions, using a 20-pixel threshold and consuming an assigned generated match only once. A hit additionally requires the target positions to differ by less than 20 pixels. $M_{\text{rate}}$ is the number of hits divided by the number of reference matches; $M_{\text{dist}}$ averages target-position distances over source-assigned matches, including those that miss the target threshold. Dataset-level scores aggregate the corresponding counts and distance sums, rather than averaging scene-level ratios.

\paragraph{Dynamic consistency and quality.}
We use VBench's custom-input evaluation for the five dimensions in~\cref{tab:dynamic}: subject consistency, background consistency, motion smoothness, aesthetic quality, and imaging quality. Generated and ground-truth videos are evaluated separately with the same implementation, and each dimension is averaged over videos. 

\subsection{Inference Cost}
The costs in~\cref{tab:dynamic} are measured on one NVIDIA A800 using three common OmniWorld-Game scenes, with methods run sequentially. The Wan variants generate 81 frames at $832\times480$ using 50 sampling steps, guidance scale 5, and seed 0. Each external method retains its native resolution, temporal sampling, precision, and offloading policy. The comparison therefore measures the cost of completing the common generation task under each method's evaluation protocol.

Runtime includes required preprocessing, generation, output postprocessing, and generated-video encoding, but excludes model loading and uninstrumented process overhead. GPU memory is monitored with NVML over the scene's process tree, with a target sampling interval of 50 ms and decimal GB units. Average memory is the time-weighted process-tree memory over the full scene-command duration, including loading and export; peak memory is the maximum sampled value in that run. The table reports the arithmetic mean of each quantity over the three scenes, so its peak-memory column is the mean of the per-scene peaks. For \modelname, the trained DiT replaces the corresponding backbone DiT, and no teacher or aggregation MLP is loaded.

\section{Additional Quantitative Results}
\label[appendix]{app:additional-ablations}

\cref{tab:timestep-routing-full,tab:duration-full} report the complete results corresponding to the compact comparisons in~\cref{tab:timestep-routing,tab:duration}, including rendering metrics and the Wan baseline (R1).

\paragraph{Loss assignment configurations.}
The Loss Assignment column in~\cref{tab:timestep-routing-full} specifies where the pairwise and trackwise auxiliary losses are applied during training. The flow-matching loss remains active in every configuration and at every noise timestep. Here, $t\in[0,1000]$ is the model noise timestep, with larger values indicating higher noise, and L$k$ denotes student layer $k$ using zero-based indexing. The eight configurations are:
\begin{itemize}
  \item \textbf{R1 (Off @ all $t$):} The Wan baseline, with no auxiliary attention supervision at any noise timestep.
  \item \textbf{R2 (L5 @ all $t$):} Both auxiliary losses are applied at layer 5 throughout the noise range, testing a fixed shallow supervision layer.
  \item \textbf{R3 (L15 @ all $t$):} Both auxiliary losses are applied at layer 15 throughout the noise range, testing a fixed intermediate supervision layer.
  \item \textbf{R4 (L25 @ all $t$):} Both auxiliary losses are applied at layer 25 throughout the noise range, testing a fixed deep supervision layer.
  \item \textbf{R5 (L5 @ $t\ge900$):} Auxiliary supervision is applied only at layer 5 for high-noise samples with $t\ge900$; samples with $t<900$ receive only the flow-matching loss.
  \item \textbf{R6 (L25 @ $t<900$):} Auxiliary supervision is applied only at layer 25 for lower-noise samples with $t<900$; samples with $t\ge900$ receive only the flow-matching loss.
  \item \textbf{R7 (L25 / L5 @ 900):} Auxiliary supervision is routed to layer 25 for $t<900$ and layer 5 for $t\ge900$. This schedule assigns a deeper layer to lower noise and a shallower layer to higher noise.
  \item \textbf{R8 (L5 / L15 / L25 @ 850, 950):} Our schedule routes auxiliary supervision to layer 5 for $t<850$, layer 15 for $850\le t<950$, and layer 25 for $t\ge950$, assigning progressively deeper layers as noise increases.
\end{itemize}
For every supervised sample, both auxiliary objectives use the same selected layer; the listed layers are not supervised simultaneously. At each routing boundary, equality belongs to the higher-noise interval. The auxiliary loss weights follow~\cref{app:training-details}.

The routed configuration improves all four geometric metrics over the tested fixed-layer and alternative schedules. These results support adapting supervision depth to the noise level. Across durations, \modelname improves the geometric metrics at 5, 10, and 15 seconds, while rendering metrics show small trade-offs at the shorter durations. Each duration uses a separately trained Wan/\modelname pair, so this comparison assesses applicability across training lengths.

\begin{table}[!htbp]
  \caption{\textbf{Complete noise-timestep routing results on DL3DV-10K with Wan2.2 5B.} R1--R8 follow~\cref{tab:timestep-routing}; R8 is \modelname. Layer sequences run from low to high noise; values after @ are model-timestep boundaries, assigned to the higher-noise interval. For inequalities, supervision is off outside the stated interval. Best results are bold.}
  \label{tab:timestep-routing-full}
  \centering
  \begingroup
  \scriptsize
  \setlength{\tabcolsep}{4pt}
  \renewcommand{\arraystretch}{1.15}
  \begin{tabular}{@{}lc*{7}{c}@{}}
    \toprule
    Exp. ID & Loss Assignment & FID$\downarrow$ & FVD$\downarrow$ & CLIP$\uparrow$ & $R_{\text{dist}}\downarrow$ & $T_{\text{dist}}\downarrow$ & $M_{\text{rate}}\uparrow$ & $M_{\text{dist}}\downarrow$ \\
    \midrule
    R1 (Wan) & Off @ all $t$ & 38.60 & 322.1 & 33.85 & 9.573 & 0.108 & 0.257 & 48.90 \\
    \midrule
    R2 & L5 @ all $t$ & 38.03 & 321.0 & \textbf{34.24} & 7.440 & 0.086 & 0.295 & 43.89 \\
    R3 & L15 @ all $t$ & 38.18 & 319.1 & 34.12 & 7.346 & 0.090 & 0.286 & 42.71 \\
    R4 & L25 @ all $t$ & 38.07 & 320.3 & 34.17 & 6.540 & 0.075 & 0.289 & 43.55 \\
    \midrule
    R5 & L5 @ $t\ge900$ & 38.64 & 323.9 & 34.09 & 8.932 & 0.098 & 0.262 & 46.36 \\
    R6 & L25 @ $t<900$ & 38.80 & 322.9 & 34.03 & 9.290 & 0.106 & 0.258 & 47.17 \\
    R7 & L25 / L5 @ 900 & 38.48 & 320.8 & 34.12 & 8.755 & 0.104 & 0.268 & 45.77 \\
    \midrule
    \textbf{R8 (ours)} & \textbf{L5 / L15 / L25 @ 850, 950} & \textbf{37.93} & \textbf{318.0} & \textbf{34.24} & \textbf{4.890} & \textbf{0.065} & \textbf{0.335} & \textbf{38.85} \\
    \bottomrule
  \end{tabular}%
  \endgroup
\end{table}

\begin{table}[!htbp]
  \caption{\textbf{Complete generation-duration results on DL3DV-10K with Wan2.2 5B.} Separately trained Wan/\modelname pairs show geometric gains at every duration, with small FID or FVD trade-offs at 5 and 10 seconds. Best results within each duration are bold.}
  \label{tab:duration-full}
  \centering
  \begingroup
  \scriptsize
  \setlength{\tabcolsep}{4pt}
  \renewcommand{\arraystretch}{1.15}
  \begin{tabular}{@{}l*{7}{c}@{}}
    \toprule
    Duration & FID$\downarrow$ & FVD$\downarrow$ & CLIP$\uparrow$ & $R_{\text{dist}}\downarrow$ & $T_{\text{dist}}\downarrow$ & $M_{\text{rate}}\uparrow$ & $M_{\text{dist}}\downarrow$ \\
    \midrule
    5s & \textbf{26.85} & 285.7 & 34.22 & 3.020 & 0.064 & 0.529 & 21.84 \\
    5s $+$\,\modelname & 27.63 & \textbf{285.1} & \textbf{34.26} & \textbf{1.684} & \textbf{0.053} & \textbf{0.571} & \textbf{19.31} \\
    \midrule
    10s & 33.68 & \textbf{287.4} & 34.03 & 5.948 & 0.072 & 0.371 & 38.34 \\
    10s $+$\,\modelname & \textbf{33.49} & 288.4 & \textbf{34.14} & \textbf{3.376} & \textbf{0.056} & \textbf{0.453} & \textbf{33.33} \\
    \midrule
    15s & 38.60 & 322.1 & 33.85 & 9.573 & 0.108 & 0.257 & 48.90 \\
    15s $+$\,\modelname & \textbf{37.93} & \textbf{318.0} & \textbf{34.24} & \textbf{4.890} & \textbf{0.065} & \textbf{0.335} & \textbf{38.85} \\
    \bottomrule
  \end{tabular}%
  \endgroup
\end{table}

\section{PCA and Attention Analysis Details}
\label[appendix]{app:pca-attention}

\subsection{Video Features and Attention}
\label[appendix]{app:video-attention}
\paragraph{Generation and feature capture.}
The analysis in~\cref{fig:noise-depth} uses Wan2.2 5B on a RealEstate10K example, generating 81 frames at $832\times480$ with 50 sampling steps, guidance scale 5, scheduler shift 5, and seed 0. Features and attention are captured from the positive classifier-free-guidance branch during the same complete generation trajectory. The source query is fixed in the first latent frame at spatial token $(x,y)=(14,8)$ on the $26\times15$ grid. The target is latent frame 10, displayed over frame 40 of the final generated video. This displayed frame is a visualization representative, distinct from the temporal representatives used to extract training labels.

\paragraph{PCA projection.}
For each inspected layer and denoising step, we collect the target frame's block-output features and project the spatial tokens onto three principal components. PCA is fitted independently for each layer--step pair, with a fixed random seed and deterministic component signs. Each projected channel is scaled using its first and 99th percentiles for RGB display. Because each projection is fitted independently, we compare the spatial organization of regions within each visualization rather than matching colors across cells. As discussed in~\cref{sec:video-attention}, these maps reveal where recognizable scene structure emerges across noise levels and network depth: little coherent structure is visible in the noise-dominated region, scene regions begin to take shape in the structure-formation region, and the scene layout becomes more clearly defined in the structure-established region. The onset of recognizable structure occurs in deeper layers at higher noise and in shallower layers at lower noise. Viewed alongside the attention maps, the PCA features help distinguish the emergence of scene structure from the development of cross-frame associations, showing that attention can still favor competing regions even when the scene layout is recognizable. This joint analysis motivates placing correspondence supervision near the structure-formation stage and adapting the supervised layer to the noise timestep.

\paragraph{Attention readout.}
At the same layer and step, we extract queries and keys after normalization and rotary positional encoding. For the fixed source query, each head's dot-product scores are softmax-normalized over all video keys. We then retain the target-frame slice and average the resulting probabilities over heads. This readout differs from both the block-output features used for PCA and the learned aggregation MLP used during training. Target-frame responses are displayed after normalization by each map's maximum, with color mapping and interpolation used only for visualization. Thus, the color scale indicates relative spatial response within a map, not an absolute probability shared across layers or noise levels. 

The denoising axis uses the actual model noise timesteps along the sampled inference trajectory. The noise-dependent shift in structure formation motivates our layer-routing schedule, whose generation performance is evaluated against fixed-layer and alternative schedules in~\cref{tab:timestep-routing-full}.

\subsection{Geometry Attention}
\label[appendix]{app:geometry-attention}
The examples in~\cref{fig:vggt-omega} inspect VGGT-$\Omega$ responses to a fixed first-frame query across target frames. The static example uses the generated indoor video from the video-attention analysis, while the dynamic example uses the bike-packing sequence from TAPVid-DAVIS. These examples provide qualitative correspondence visualizations. The teacher receives the complete clean input sequence: 81 frames for the static example and 69 frames for the dynamic example.

For training-label extraction, we use the fixed global-attention layer 13, head 10 (zero-based indices), as described in~\cref{app:teacher-details}. For these visualizations, source queries are mapped to the $26\times15$ grid (width $\times$ height), and target patch responses are resampled and normalized within each target frame. The static query lies on the stair railing; the dynamic query lies on the person's hat. Target frames are selected across the sequence for visualization.

Heatmaps are normalized within each target frame and displayed with contrast adjustment to highlight the spatial concentration of attention responses. The examples illustrate that selected heads can associate observations of static surfaces and moving object parts.

\section{Limitations and Future Work}
\label[appendix]{app:limitations}

\modelname improves scene and dynamic consistency across all three video backbones by transferring geometric correspondences during training, while preserving the original inference procedure. Its current design suggests three directions for further improvement.

\paragraph{More reliable correspondence targets.}
Our results demonstrate that attention distillation with geometric correspondence targets improves video consistency. The current implementation extracts these targets from VGGT-$\Omega$ attention, making their quality dependent on the teacher's matching ability and spatial resolution. This raises a natural question: can more reliable correspondences further improve the effectiveness of attention distillation? Stronger correspondence estimators or complementary sources of matching evidence could provide more accurate and complete supervision, particularly for ambiguous appearance, occlusion, and moving objects. The correspondence-based formulation of \modelname provides a basis for exploring such alternatives to the current geometry teacher.

\paragraph{Adaptive supervision placement.}
Our noise-timestep routing improves geometric metrics over fixed-layer supervision, demonstrating the value of adapting supervision depth to the denoising state. The current layer assignment depends only on the noise timestep and remains fixed throughout training. As the video model learns, the layers that benefit most from correspondence supervision may also change. Conditioning routing jointly on noise timestep and training progress could better match supervision to the model's evolving representations, extending the current mechanism while preserving the original inference procedure.

\paragraph{Extension to autoregressive video generation.}
\modelname improves static and dynamic consistency in Wan's bidirectional video models. Autoregressive video generation offers a natural next setting, where preserving scene identity over long rollouts and maintaining coherent object motion remain important challenges. The principle of guiding attention with correspondences also applies to associations between newly generated content and preceding frames. Adapting the distillation objectives to the available temporal context under causal attention could help autoregressive models retain scene structure and moving-object identity as generation proceeds. This offers a promising route to extending the benefits demonstrated here to autoregressive generation.

\section{More Qualitative Results and Videos}
\label[appendix]{app:more-results}

\begin{figure}[t]
  \centering
  \includegraphics[width=\linewidth]{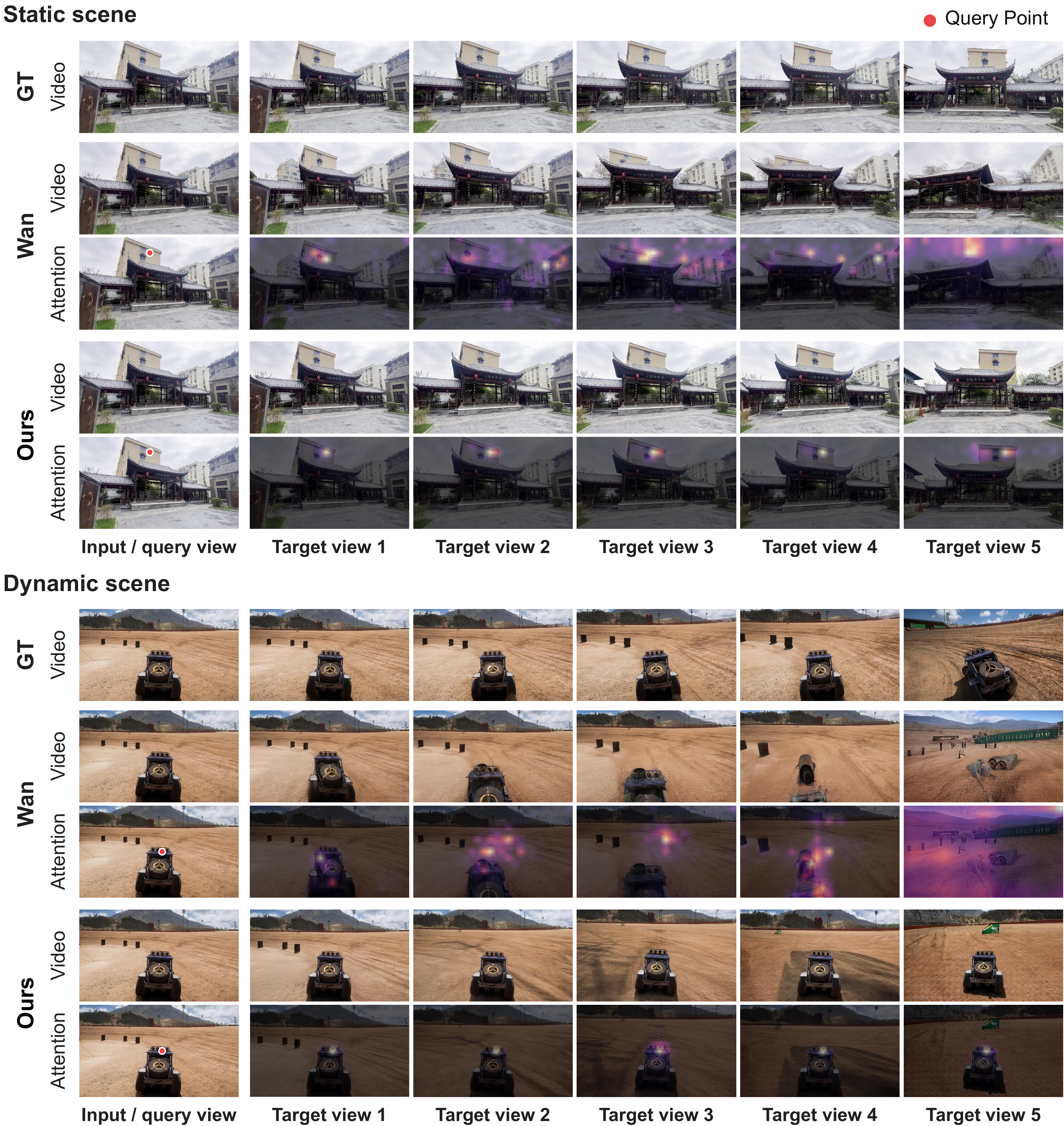}
  \caption{\textbf{Video attention and generation results with geometry attention distillation.} Compared with the Wan baseline, \modelname produces more localized attention responses around corresponding regions and better preserves scene and object structure in static (top) and dynamic (bottom) examples. Red dots mark source queries in the input frame; heatmaps show their attention responses in subsequent target views. GT denotes reference frames.}
  \label{fig:attention_comparison}
\end{figure}

\begin{figure}[t]
  \centering
  \includegraphics[width=\linewidth]{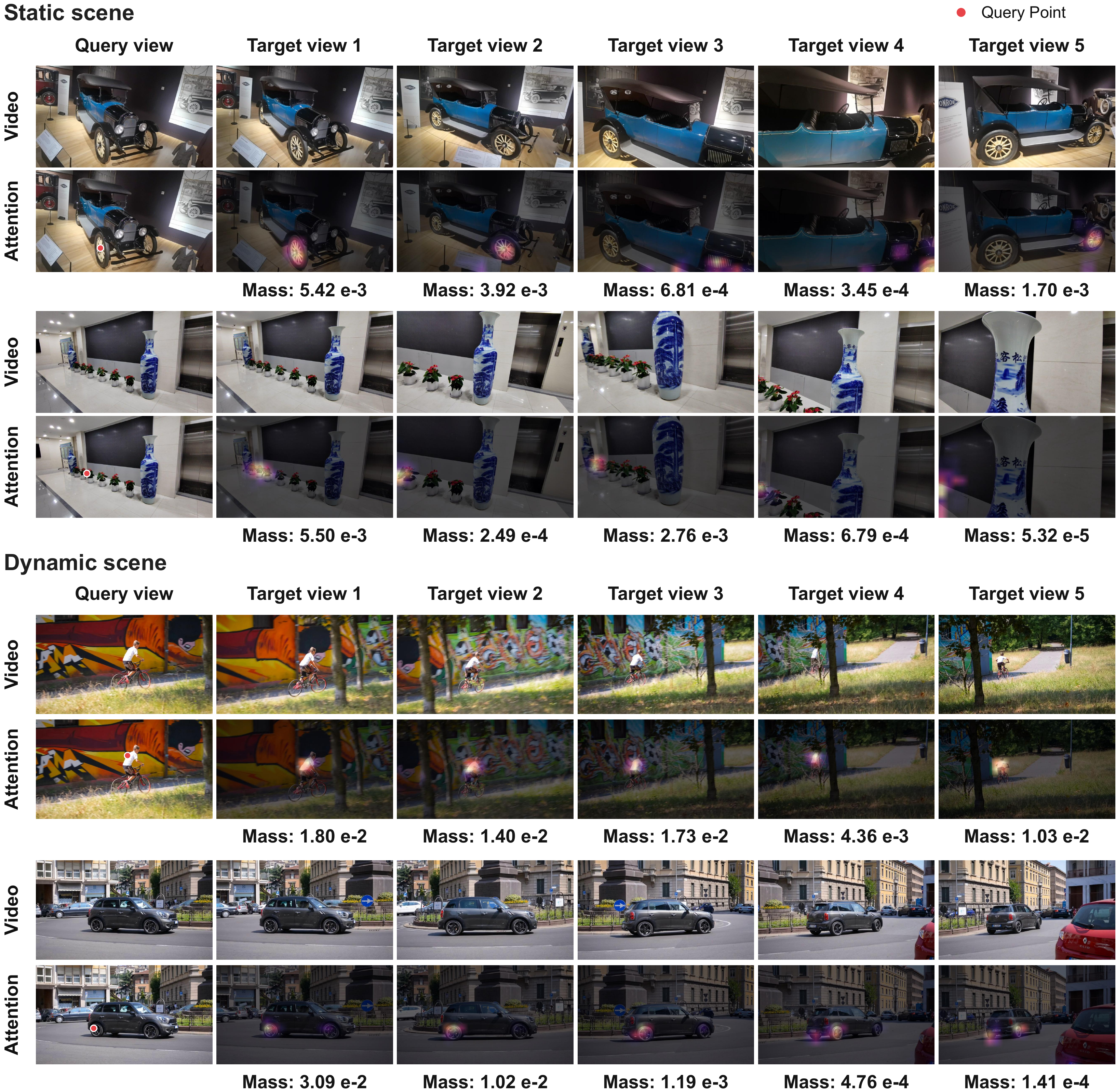}
  \caption{\textbf{VGGT-$\Omega$ attention and correspondence reliability.} Static (top) and dynamic (bottom) examples show that ambiguous or incorrect responses can have low attention mass despite visible peaks, supporting its use as a reliability cue. Red dots mark source queries; heatmaps show target-view responses, with the total attention mass assigned to each target frame reported below.}
  \label{fig:more_vggt_omega}
\end{figure}

\begin{figure}[t]
  \centering
  \includegraphics[width=0.92\linewidth]{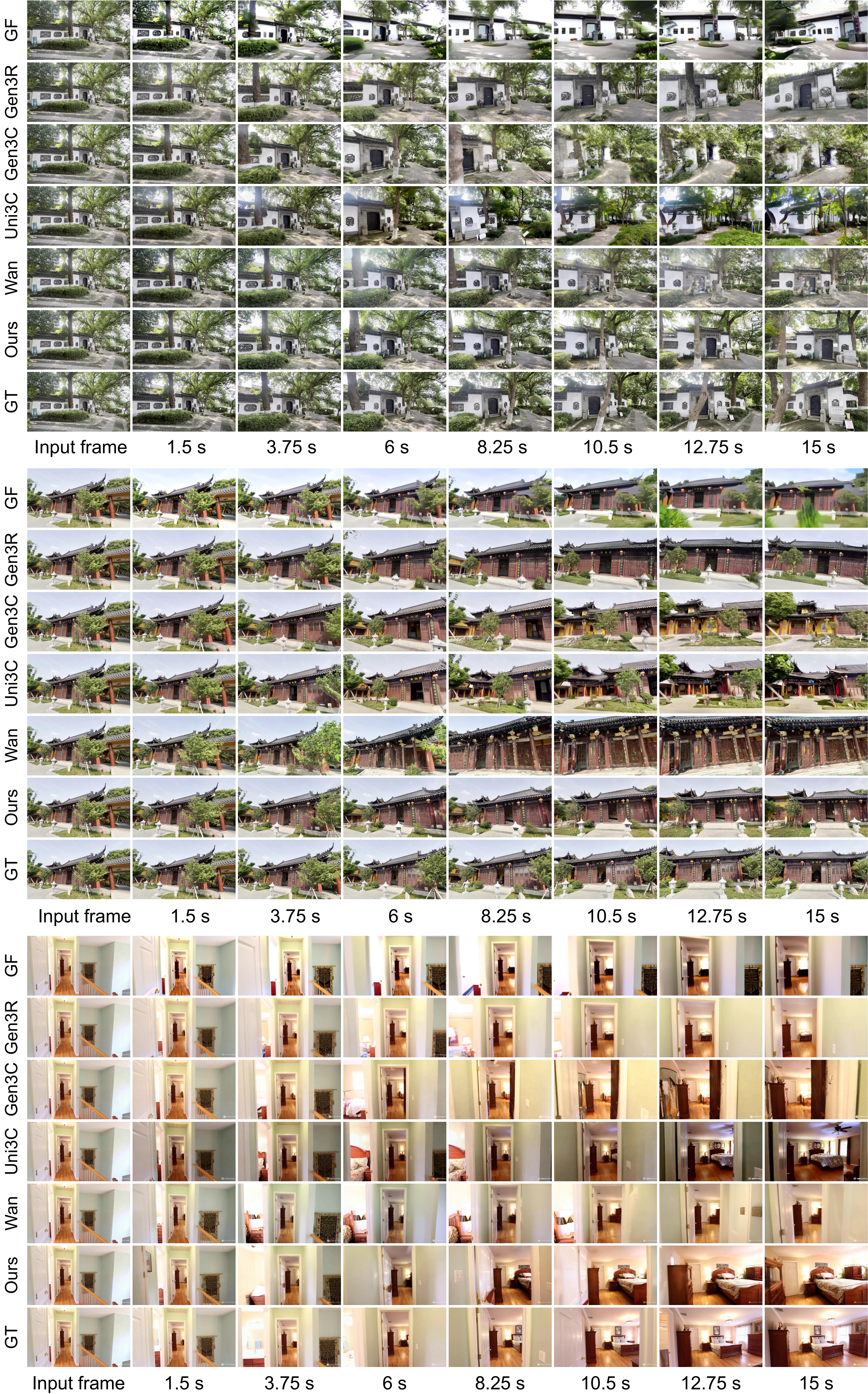}
  \caption{\textbf{More 15-second camera-controlled generation results in static scenes.}}
  \label{fig:qual_static_results_s1}
\end{figure}

\begin{figure}[t]
  \centering
  \includegraphics[width=0.92\linewidth]{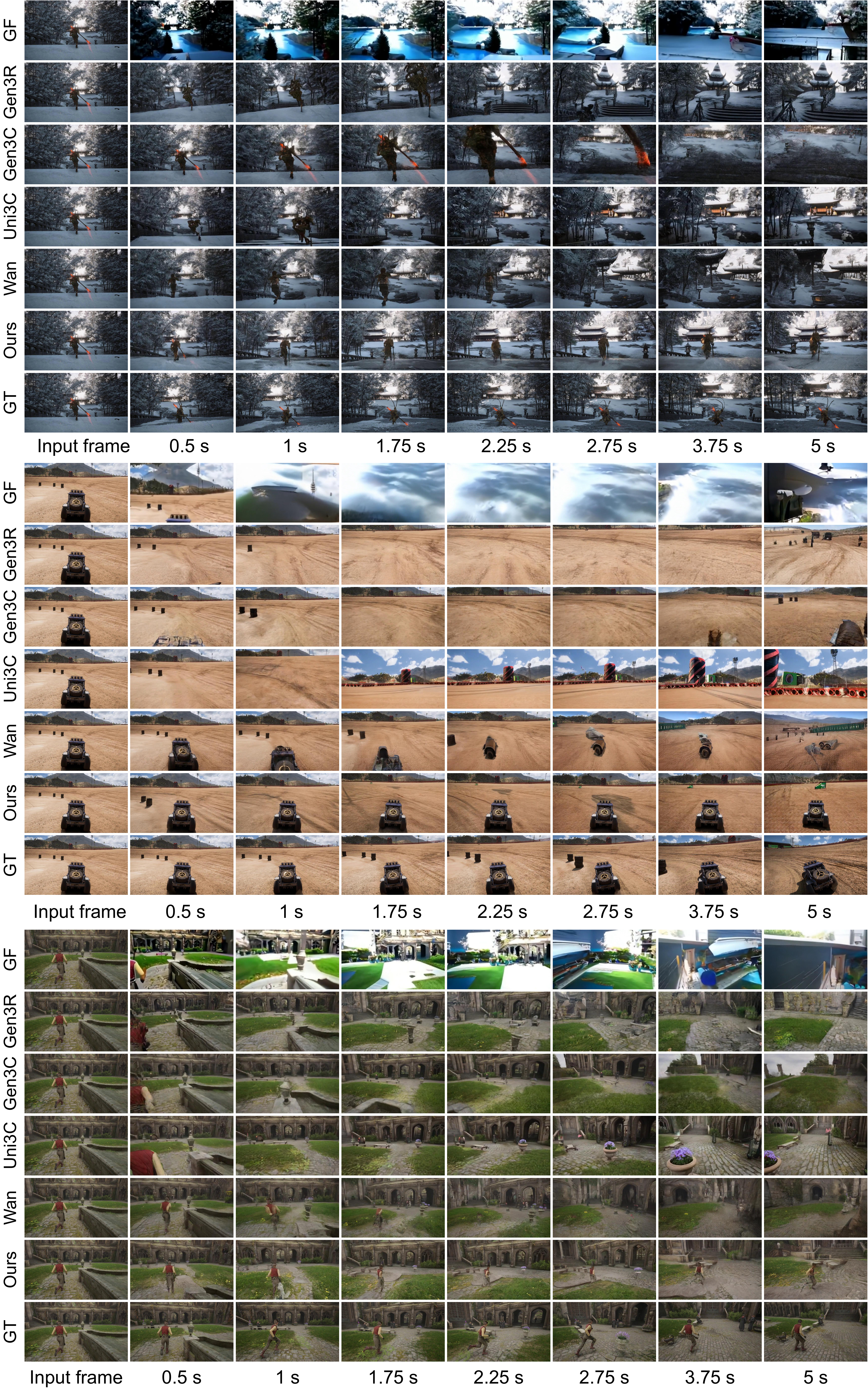}
  \caption{\textbf{More 5-second camera-controlled generation results in dynamic scenes.}}
  \label{fig:qual_dynamic_results_s2}
\end{figure}

\cref{fig:attention_comparison,fig:more_vggt_omega} provide additional visualizations of video attention with \modelname and VGGT-$\Omega$ correspondence reliability, respectively. \cref{fig:qual_static_results_s1,fig:qual_dynamic_results_s2} provide additional qualitative comparisons for long-range static scenes and dynamic scenes, respectively. All examples use first-frame, text, and camera conditioning. 

\end{document}

%% file: math_commands.tex
\usepackage{amsmath,amsfonts,bm}

\def\eqref#1{equation~\ref{#1}}

\def\1{\bm{1}}

\DeclareMathAlphabet{\mathsfit}{\encodingdefault}{\sfdefault}{m}{sl}
\SetMathAlphabet{\mathsfit}{bold}{\encodingdefault}{\sfdefault}{bx}{n}

\newcommand{\softmax}{\mathrm{softmax}}
\newcommand{\sigmoid}{\sigma}



%% file: config.tex
\usepackage[utf8]{inputenc} 
\usepackage[T1]{fontenc}    
\usepackage{graphicx}
\usepackage{amsmath}
\usepackage{amssymb}
\usepackage{booktabs}
\usepackage{enumitem}
\usepackage{colortbl} 
\usepackage{acronym}
\usepackage{balance}
\usepackage{xspace}
\usepackage{setspace}
\usepackage{tabularx,colortbl,multirow,array,makecell}
\usepackage{overpic,wrapfig}
\usepackage{nicefrac}
\usepackage[misc]{ifsym}
\usepackage{siunitx}
\usepackage[dvipsnames,svgnames,x11names]{xcolor}
\usepackage{pifont}
\usepackage{dsfont}
\usepackage[capitalize]{cleveref}
\usepackage{bm}
\usepackage{caption}
\usepackage{blindtext}
\usepackage{subcaption}

\definecolor{gblue}{HTML}{4285F4}
\definecolor{gred}{HTML}{DB4437}
\definecolor{ggreen}{HTML}{0F9D58}
\definecolor{vblue}{HTML}{2993ba}

\definecolor{gbest}{HTML}{FFCCCB}
\definecolor{gsecond}{HTML}{FFE5CC}
\definecolor{gthird}{HTML}{FFF2A0}

\acrodef{mlp}[MLP]{multi-layer perceptron}
\acrodef{gt}[GT]{Ground Truth}
\acrodef{miou}[mIoU]{Mean Intersection over Union}
\acrodef{fr}[FR]{full-reference}
\acrodef{nr}[NR]{no-reference}

\newcommand{\modelname}{\textbf{\textsc{GAD}}\xspace}

\def\onedot{.\xspace}
 
\def\ie{\emph{i.e}\onedot}

\renewcommand{\sigmoid}{\operatorname{sigmoid}}